%% file: main.tex
\documentclass[acmsmall,screen]{acmart}

\usepackage{array}
\usepackage{tabularx}

\newif\ifdynamicfigures
\dynamicfigurestrue
\ifdynamicfigures
  \usepackage{tikz}
  \usetikzlibrary{arrows.meta,calc}
  \definecolor{cFound}{HTML}{2C5F8A}
  \definecolor{cData}{HTML}{B0453C}
  \definecolor{cArch}{HTML}{C8791F}
  \definecolor{cTrain}{HTML}{6A4A9C}
  \definecolor{cHybrid}{HTML}{2F7D55}
  \definecolor{cTool}{HTML}{8A4A93}
  \definecolor{cEval}{HTML}{1F7A8C}
  \definecolor{cApp}{HTML}{B03A63}
  \definecolor{cInk}{HTML}{1C2126}
  \definecolor{cMute}{HTML}{737C85}
  \definecolor{cPaper}{HTML}{FBFBFC}
  \newcommand{\figfont}{\sffamily}
  \tikzset{
    axisline/.style={
      -{Stealth[length=9pt,width=7pt]}, draw=black, line width=1.8pt},
  }
\fi

\makeatletter
\def\@secfont{\sffamily\bfseries\large\section@raggedright}
\def\@subsecfont{\sffamily\bfseries\section@raggedright}
\def\@subsubsecfont{\sffamily\bfseries\itshape}
\def\@parfont{\itshape}
\renewcommand\section{\def\@toclevel{1}%
  \@startsection{section}{1}{\z@}%
  {-1.05\baselineskip \@plus -2\p@ \@minus -.2\p@}%
  {.3\baselineskip}%
  {\ACM@NRadjust\@secfont}}
\renewcommand\subsection{\def\@toclevel{2}%
  \@startsection{subsection}{2}{\z@}%
  {-.75\baselineskip \@plus -2\p@ \@minus -.2\p@}%
  {.15\baselineskip}%
  {\ACM@NRadjust\@subsecfont}}
\renewcommand\subsubsection{\def\@toclevel{3}%
  \@startsection{subsubsection}{3}{\z@}%
  {-.5\baselineskip \@plus -2\p@ \@minus -.2\p@}%
  {-3.5\p@}%
  {\ACM@NRadjust{\@subsubsecfont\@adddotafter}}}
\renewcommand\paragraph{\def\@toclevel{4}%
  \@startsection{paragraph}{4}{\parindent}%
  {-.4\baselineskip \@plus -2\p@ \@minus -.2\p@}%
  {-3.5\p@}%
  {\ACM@NRadjust{\@parfont\@adddotafter}}}

\let\ACM@origsection\section
\let\ACM@origsubsection\subsection
\let\ACM@origsubsubsection\subsubsection
\let\ACM@origparagraph\paragraph
\makeatother

\newcolumntype{C}{>{\centering\arraybackslash}X}
\newcolumntype{L}{>{\raggedright\arraybackslash}X}
\newcolumntype{W}[1]{>{\hsize=#1\hsize}C}
\renewcommand{\arraystretch}{1.05}

\acmJournal{CSUR}
\acmVolume{0}
\acmNumber{0}
\acmArticle{0}
\acmMonth{1}
\acmYear{2026}
\acmDOI{XXXXXXX.XXXXXXX}

\setcopyright{acmlicensed}
\copyrightyear{2026}

\begin{document}

\title{LLM-based Agents for Forecasting and Prediction: Methods, Training, Evaluation, and Applications}

\author{Xiaogang Xu}
\affiliation{%
  \institution{Zhejiang University}
  \country{China}}

\author{Jiaqi Tang}
\correspondingauthor
\affiliation{%
  \institution{The Hong Kong University of Science and Technology}
  \country{Hong Kong}}

\author{Jianmin Chen}
\author{Yingying Yan}
\affiliation{%
  \institution{Northwestern Polytechnical University}
  \country{China}}

\author{Zhenchao Tang}
\author{Xiangxin Zhou}
\affiliation{%
  \institution{Tencent}
  \country{China}}

\author{Xiaobin Hu}
\affiliation{%
  \institution{National University of Singapore}
  \country{Singapore}}

\author{Wei Wei}
\affiliation{%
  \institution{Northwestern Polytechnical University}
  \country{China}}

\author{Jinfeng Wu}
\author{Qifeng Chen}
\affiliation{%
  \institution{The Hong Kong University of Science and Technology}
  \country{Hong Kong}}

\author{Lu Zhou}
\affiliation{%
  \institution{Nanjing University of Aeronautics and Astronautics}
  \country{China}}

\author{Jiafei Wu}
\author{Zhe Liu}
\correspondingauthor
\author{Jianwei Yin}
\affiliation{%
  \institution{Zhejiang University}
  \country{China}}

\author{Weimin Zheng}
\affiliation{%
  \institution{Tsinghua University}
  \country{China}}

\thanks{\textsuperscript{$\dagger$}Corresponding to Jiaqi Tang (\url{jtang092@connect.ust.hk}) and Zhe Liu (\url{zhe.liu@zju.edu.cn}).}%

\renewcommand{\shortauthors}{Xu et al.}

\begin{abstract}
Large language models (LLMs) now support forecasting systems that combine language-based reasoning with temporal data, evidence retrieval, external tools, and iterative prediction. We investigate LLM-based forecasting agents, meaning systems in which a language model contributes to a scored prediction about a future or currently unobserved target. We organize architectures into three groups. Standalone LLM workflows operate on encoded time series or event context. Tool- and retrieval-augmented agents incorporate external evidence. Hybrid systems pair LLMs with statistical or foundation models. We then review training methods and evaluation protocols. We examine negative as well as positive evidence, including sensitivity to small input perturbations, ablations in which the LLM component does not improve accuracy, and benchmark gains that may reflect contamination instead of temporal reasoning. We cover applications in finance, weather, health, energy, and operations, and we summarize the benchmarks and datasets used for evaluation. The evidence indicates that measurement is a central limitation. Future work requires calibration under distribution shift, contamination-resistant live evaluation, explicit reporting of cost and accuracy together, and methods for handling feedback between deployed forecasts and the outcomes being forecast.
\end{abstract}



\maketitle

\section{Introduction}
\input{sections/01_introduction}

\section{Scope and Formulation}\label{sec:methodology}
\input{sections/02_methodology}

\section{Agent Architectures}\label{sec:architectures}
\input{sections/04_data-prompting}

\section{Training and Adaptation}\label{sec:training}
\input{sections/08_training}

\section{Evaluation}\label{sec:evaluation}
\input{sections/09_benchmarks}

\section{Applications}\label{sec:applications}
\input{sections/11_applications}

\section{Risks and Open Problems}\label{sec:challenges}
\input{sections/12_risks}

\section{Conclusion}\label{sec:conclusion}
\input{sections/14_conclusion}

\bibliographystyle{ACM-Reference-Format}
\bibliography{refs}

\end{document}

%% file: sections/01_introduction.tex
Forecasting supports decisions in finance~\citep{makridakis2024m6,sezer2020financial,ryll2019evaluating}, public health~\citep{rodriguez2022data,cramer2022evaluation,reich2019collaborative}, energy~\citep{hong2016probabilistic,weron2014electricity}, operations~\citep{fildes2008forecasting,makridakis2022m5}, and climate response~\citep{lam2023graphcast,pathak2022fourcastnet,bi2022pangu}. For decades the field relied on statistical and econometric models~\citep{hyndman2008automatic}, and later on deep networks trained on structured numerical data~\citep{lim2021time,benidis2022deep}. Large language models (LLMs)~\citep{brown2020language,touvron2023llama} add semantic representations and capabilities for reasoning over events,policies, and explanatory contexts. Current evidence does not establish that LLMs can extrapolate numerical sequences more accurately than specialized forecasting models. In an agentic system, however, an LLM can retrieve evidence, invoke tools, and revise a forecast through multiple inference steps (Figure~\ref{fig:pipeline}). These capabilities come with substantial trade-offs. Forecasts that rely on unstructured evidence, such as news, filings, or health reports, may fail to improve accuracy or calibration over specialized models while requiring substantially more inference compute. One ablation study found that removing the language model component from several LLM-based forecasters left accuracy unchanged or improved it~\citep{tan2024actually}. We thus focus on the conditions under which a language-based component improves a forecast enough to justify its cost.

We study systems in which an LLM contributes, through reasoning, representation, or control, to a scored prediction about a future target. 
The review follows how such systems are formulated, built, trained, evaluated, and deployed. Some operate as prompt-only workflows over encoded series or event context. Others retrieve documents, call tools, or iterate through a reason-and-act loop. Hybrid designs pair an LLM with a statistical or time-series foundation model and assign the language component a narrower role. The rest of the paper asks when that component is worth its cost, and which evaluation controls make a reported gain credible.

Early prompt-only work treated frozen LLMs as zero-shot forecasters over serialized numbers~\citep{gruver2023llmtime}. Later methods adapt language backbones to temporal data through reprogramming, alignment, or prompt learning~\citep{zhou2023onefitsall,jin2024timellm,cao2024tempo}, and they are now compared against purpose-built time-series foundation
models~\citep{ansari2024chronos,das2024timesfm,woo2024moirai}. Other systems assign probabilities to event questions, a line of work that grew from neural forecasting of world events~\citep{zou2022autocast}. Retrieval-and-reasoning methods have approached human-forecaster accuracy on selected question sets~\citep{halawi2024approaching}, and live benchmarks compare these systems with experts and prediction markets while reducing contamination risk~\citep{karger2024forecastbench,zeng2025futurex}. The same evaluation challenges recur across these designs, including lookahead leakage, pre-training contamination, and incomplete inference-cost reporting. These limitations also affect domain deployments, where reliability depends on controlling information availability, evaluation protocols, and computational costs.

\input{figures/fig_pipeline}

Language-based evidence is particularly valuable when predictive signals are not fully represented in historical numerical observations. Policy announcements, filings, news, and outbreak reports may change an outlook before their effects appear in a numerical series~\citep{williams2024context}. An agent can retrieve timestamped evidence, call a specialist model or calculator, and preserve intermediate state as information arrives. These functions distinguish agentic forecasting from one-shot prompting, but they also introduce decisions about source validity, tool selection, stopping, and inference cost~\citep{cheng2026model}. Greater agency thus enables adaptation without by itself demonstrating higher forecast accuracy.

Our operational boundary requires an LLM to contribute to a verifiable, explicitly scored prediction. Eligible outputs include points, predictive distributions, categories, and event probabilities, and prompt-only systems remain the least agentic endpoint of the comparison. We exclude narrative commentary without a scorable forecast, purely numerical transformers except as baselines, trajectory and motion prediction~\citep{xu2025trajectory}, retrospective causal inference, and general agent benchmarks without prediction tasks. This boundary covers both numerical and judgmental forecasting while keeping outcome-based evaluation common across architectures.

Application constraints differ: finance evaluates returns as well as forecast scores, weather and health have strong specialist baselines, and energy and operations impose latency and integration limits (Section~\ref{sec:applications}). Figure~\ref{fig:corpus} groups the living corpus by application domain, placing general methods, benchmarks, and foundations in Cross-domain. These domains also expose different failure costs: a poorly calibrated probability, delayed warning, and unprofitable trading action cannot be compared through point error alone.

Prior papers cover LLM time-series architectures~\citep{zhang2024large}, trajectory and financial prediction~\citep{xu2025trajectory,vuong2024bibliometric}, or general data-science agents~\citep{rahman2025llm,zhou2025survey}. Others
focus on temporal reasoning, post-training, agent frameworks, multimodal fusion, or pre-agentic models~\citep{chang2025reasoningsurvey,xie2026posttraining,
cheng2026agentic,jiang2025multimodalsurvey,shi2025llmtssurvey}.Existing architecture papers generally place less emphasis on tool-mediated control, live evidence, and outcome-based evaluation, while broader agent papers typically treat forecasting as one task among many. Reviews of temporal reasoning or post-training cover only parts of the forecasting pipeline. Our narrower, operational scope jointly examines numerical and event targets from formulation through deployment, including component ablations, calibration, contamination controls, inference costs (Section~\ref{sec:metrics}), and deployment risks (Section~\ref{sec:challenges}). Positive and negative results are judged under the same evidentiary standard.

\vspace{-0.1in} \paragraph{Contributions}
This study makes five contributions. First, we define the scope of LLM-based forecasting agents and place prompt-only models and tool-using systems on a single continuum of agency (Sections~\ref{ssec:formulation} and~\ref{ssec:agent-def}). Second, we provide a taxonomy spanning data representations, agent architectures, training paradigms, hybrid models, and tool augmentation (Figure~\ref{fig:taxonomy}). Third, we analyze evaluation challenges, including contamination, temporal leakage, aggregation, and comparisons with human forecasters (Section~\ref{sec:evaluation}). Fourth, we examine how domain-specific constraints shape agent design across finance, weather, health, energy, and operations (Section~\ref{sec:applications}). Fifth, we derive a research agenda from current measurement and deployment limitations (Section~\ref{sec:challenges}), and we apply a uniform evidentiary standard to both positive and negative results throughout.

\input{figures/fig_taxonomy}

\vspace{-0.1in} \paragraph{Organization}
Section~\ref{sec:methodology} defines scope, tasks, agency, and baselines; Sections~\ref{sec:architectures} and~\ref{sec:training} cover architectures and training. Section~\ref{sec:evaluation} examines benchmarks, leakage, uncertainty, aggregation, and cost. Sections~\ref{sec:applications} and~\ref{sec:challenges} address deployments, risks, and open problems, and Section~\ref{sec:conclusion} concludes.

%% file: figures/fig_pipeline.tex
\begin{figure}[t]
\centering
\includegraphics[width=\textwidth]{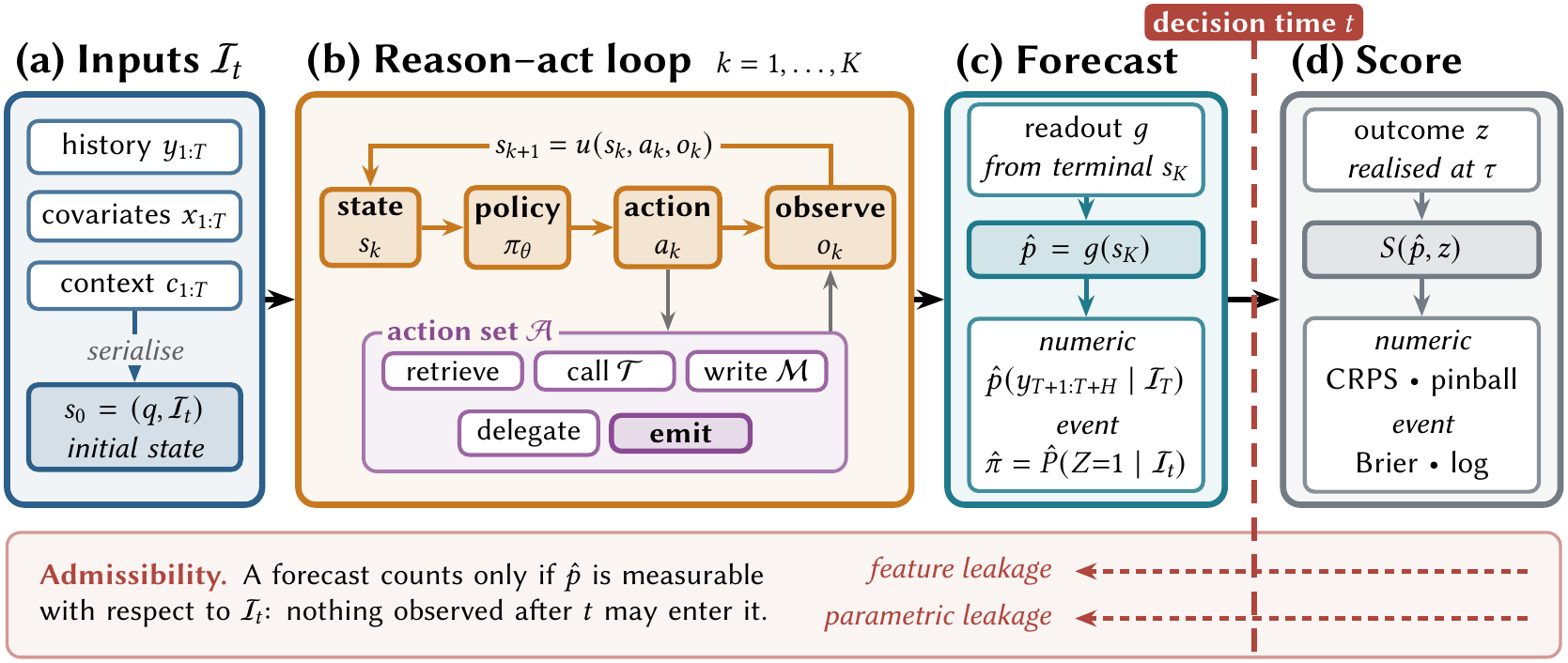}
\caption{Forecasting-agent pipeline corresponding to the formalism in
Sections~\ref{ssec:formulation}--\ref{ssec:agent-def}. Starting from the
information set $\mathcal{I}_t$, the agent executes a reason--act loop and uses
the readout $g$ to produce a forecast. An appropriate forecast score evaluates the forecast when the outcome is
observed at $\tau$. The dashed line marks decision
time $t$, and the backward arrows show feature and parametric leakage across
that boundary. Setting $K = 1$ and
$\mathcal{A} = \{\textsc{emit}\}$ yields a prompt-only zero-shot forecaster.}
\label{fig:pipeline}
\end{figure}

%% file: figures/fig_taxonomy.tex
\begin{figure}[t]
\centering
\ifdynamicfigures
  \begingroup
  \resizebox{\textwidth}{!}{%
    \input{figures/src/taxonomy.tex}%
  }
  \endgroup
\else
  \includegraphics[width=\textwidth]{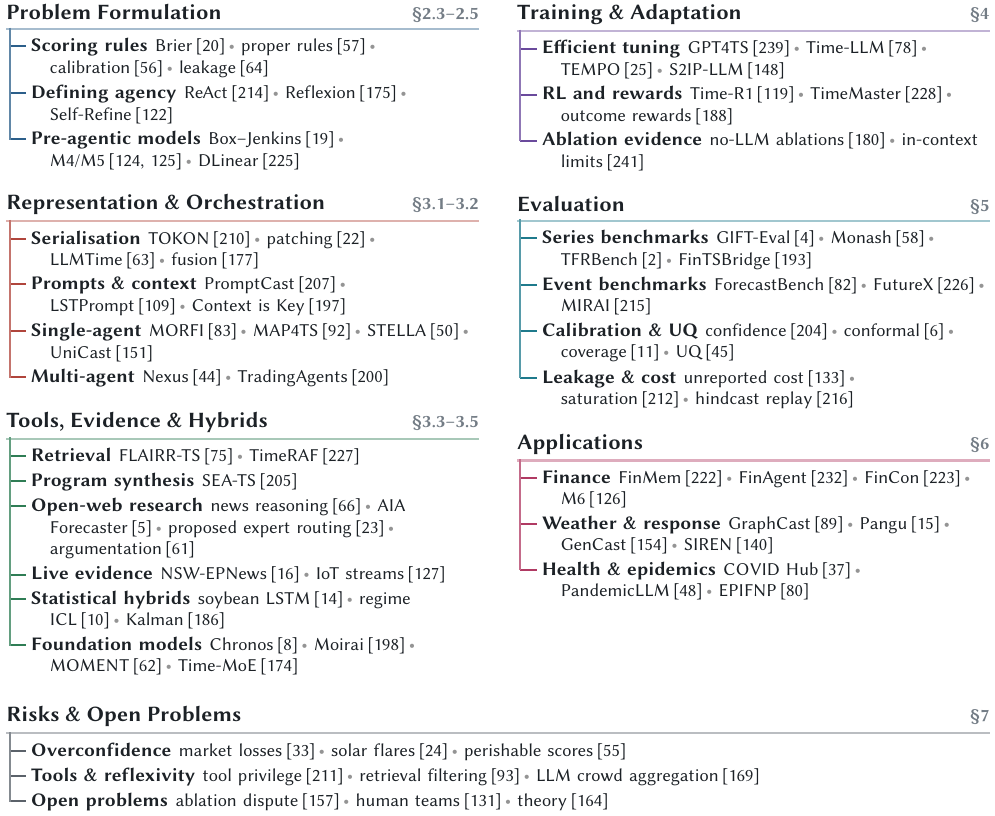}
\fi
\caption{Taxonomy of LLM-based agents for forecasting and prediction. Each
branch identifies representative work and the part of the paper that discusses
it. The first branch covers formulation, agency, and forecasting baselines in
Section~\ref{sec:methodology}. The architecture branches distinguish systems
that operate on an LLM's internal representation from systems that connect the
LLM to tools, external evidence, or other models. The final branch summarizes
risks and open problems that apply across all method families.}
\label{fig:taxonomy}
\end{figure}

%% file: figures/src/taxonomy.tex
\colorlet{cRisk}{cMute!72!cInk}%
\colorlet{cX}{cInk}%

\newcommand{\bul}{\,\textcolor{black!42}{\textbullet}\ }%
\newcommand{\nohyph}{\hyphenpenalty=10000\relax\exhyphenpenalty=10000\relax}%
\newcommand{\sct}[1]{{\fontsize{8.8}{9.8}\selectfont\S #1}}%

\newcommand{\hdrf}{\figfont\bfseries\fontsize{10.6}{11.8}\selectfont}%
\newcommand{\tagf}{\figfont\fontsize{8.8}{9.8}\selectfont\color{cMute}}%
\newcommand{\lblf}{\figfont\bfseries\fontsize{9.3}{10.4}\selectfont}%
\newcommand{\sysf}{\figfont\fontsize{8.8}{10.4}\selectfont}%

\newcommand{\txcolw}{8.00cm}%
\newcommand{\txfullw}{16.65cm}%
\newcommand{\txsubw}{206.6pt}%
\newcommand{\txbsubw}{452.7pt}%
\newcommand{\txnx}{21pt}%
\newcommand{\txspx}{-19.4pt}%

\tikzset{
  bh/.style={anchor=north west, inner sep=0pt, outer sep=0pt, text=cInk},
  br/.style={anchor=north west, inner sep=0pt, outer sep=0pt,
             text width=\txsubw, align=left, text=cInk, font=\sysf\nohyph},
  bq/.style={br, text width=\txbsubw},
}%

\newcommand{\tick}{\makebox[0pt][r]{%
  \textcolor{cX}{\rule[0.52ex]{7.8pt}{0.95pt}}\hspace{2.6pt}}}%

\newlength{\txnw}%
\newcommand{\bhdr}[5]{%
  \settowidth{\txnw}{\hdrf#3}%
  \node[bh] (#1) at #2 {\makebox[#5][s]{\hdrf#3\hfill{\tagf#4}}};%
  \draw[cX!42, line width=1.0pt]
    ([xshift=\txnw, yshift=-3.0pt]#1.south west) --
    ([xshift=#5, yshift=-3.0pt]#1.south west);%
  \draw[cX, line width=2.1pt]
    ([yshift=-3.0pt]#1.south west) -- ([xshift=\txnw, yshift=-3.0pt]#1.south west);%
}%
\newcommand{\brow}[4]{%
  \node[br] (#1) at #2 {\hspace*{-9pt}{\lblf\tick #3}\hspace{3.6pt}{\sysf #4}};%
}%
\newcommand{\qrow}[4]{%
  \node[bq] (#1) at #2 {\hspace*{-9pt}{\lblf\tick #3}\hspace{3.6pt}{\sysf #4}};%
}%
\newcommand{\bspine}[2]{%
  \draw[cX!75, line width=0.95pt]
    ([xshift=\txspx, yshift=5.2pt]#1.north west) --
    ([xshift=\txspx, yshift=-4.2pt]#2.north west);%
}%

\begin{tikzpicture}
  \coordinate (o) at (0,0);
  \colorlet{cX}{cFound}
  \bhdr{h1}{(0,0)}{Problem Formulation}%
       {\sct{\ref{ssec:formulation}--\ref{ssec:baseline}}}{\txcolw}
  \brow{a1}{([xshift=\txnx, yshift=-7.6pt]h1.south west)}{Scoring rules}%
       {Brier\,\cite{brier1950verification}\bul proper rules\,\cite{gneiting2007strictly}%
        \bul calibration\,\cite{gneiting2007probabilistic}\bul leakage\,\cite{guan2026leakage}}
  \brow{a2}{([yshift=-3.6pt]a1.south west)}{Defining agency}%
       {ReAct\,\cite{yao2023react}\bul Reflexion\,\cite{shinn2023reflexion}%
        \bul Self-Refine\,\cite{madaan2023selfrefine}}
  \brow{a3}{([yshift=-3.6pt]a2.south west)}{Pre-agentic models}%
       {Box--Jenkins\,\cite{box2015time}\bul M4/M5\,\mbox{\cite{makridakis2020m4,makridakis2022m5}}%
        \bul DLinear\,\cite{zeng2023dlinear}}
  \bspine{a1}{a3}
  \colorlet{cX}{cData}
  \bhdr{h2}{([xshift=-\txnx, yshift=-12pt]a3.south west)}{Representation \& Orchestration}%
       {\sct{\ref{ssec:tokenisation}--\ref{ssec:refinement}}}{\txcolw}
  \brow{b1}{([xshift=\txnx, yshift=-7.6pt]h2.south west)}{Serialisation}%
       {TOKON\,\cite{yang2025tokon}\bul patching\,\cite{bumb2025forecasting}%
        \bul LLMTime\,\cite{gruver2023llmtime}\bul fusion\,\cite{su2025fusing}}
  \brow{b2}{([yshift=-3.6pt]b1.south west)}{Prompts \& context}%
       {PromptCast\,\cite{xue2023promptcast}\bul LSTPrompt\,\cite{liu2024lstprompt}%
        \bul Context is Key\,\cite{williams2024context}}
  \brow{b3}{([yshift=-3.6pt]b2.south west)}{Single-agent}%
       {MORFI\,\cite{khezresmaeilzadeh2025morfi}\bul MAP4TS\,\cite{lee2025map}%
        \bul STELLA\,\cite{fan2025stella}\bul UniCast\,\cite{park2025unicast}}
  \brow{b4}{([yshift=-3.6pt]b3.south west)}{Multi-agent}%
       {Nexus\,\cite{das2026nexus}\bul TradingAgents\,\cite{xiao2024tradingagents}}
  \bspine{b1}{b4}
  \colorlet{cX}{cHybrid}
  \bhdr{h3}{([xshift=-\txnx, yshift=-12pt]b4.south west)}{Tools, Evidence \& Hybrids}%
       {\sct{\ref{ssec:tool-augmented}--\ref{ssec:tsfm}}}{\txcolw}
  \brow{c1}{([xshift=\txnx, yshift=-7.6pt]h3.south west)}{Retrieval}%
       {FLAIRR-TS\,\cite{jalori2025flairr}\bul TimeRAF\,\cite{zhang2025timeraf}}
  \brow{c2}{([yshift=-3.6pt]c1.south west)}{Program synthesis}%
       {SEA-TS\,\cite{xu2026seats}}
  \brow{c3}{([yshift=-3.6pt]c2.south west)}{Open-web research}%
       {news reasoning\,\cite{halawi2024approaching}\bul AIA Forecaster\,\cite{alur2025aiaforecaster}%
        \bul proposed expert routing\,\cite{cai2026forecastagentsearch}%
        \bul argumentation\,\cite{gorur2025retrieval}}
  \brow{c4}{([yshift=-3.6pt]c3.south west)}{Live evidence}%
       {NSW-EPNews\,\cite{bi2025nsw}\bul IoT streams\,\cite{manjunath2026llm}}
  \brow{c5}{([yshift=-3.6pt]c4.south west)}{Statistical hybrids}%
       {soybean LSTM\,\cite{benjamim2026soybean}%
        \bul regime ICL\,\cite{asaad2026regime}\bul Kalman\,\cite{tian2026regime}}
  \brow{c6}{([yshift=-3.6pt]c5.south west)}{Foundation models}%
       {Chronos\,\cite{ansari2024chronos}\bul Moirai\,\cite{woo2024moirai}%
        \bul MOMENT\,\cite{goswami2024moment}\bul Time-MoE\,\cite{shi2025timemoe}}
  \bspine{c1}{c6}
  \colorlet{cX}{cTrain}
  \bhdr{h4}{(8.65cm,0)}{Training \& Adaptation}{\sct{\ref{sec:training}}}{\txcolw}
  \brow{d1}{([xshift=\txnx, yshift=-7.6pt]h4.south west)}{Efficient tuning}%
       {GPT4TS\,\cite{zhou2023onefitsall}\bul Time-LLM\,\cite{jin2024timellm}%
        \bul TEMPO\,\cite{cao2024tempo}\bul S2IP-LLM\,\cite{pan2024s2ipllm}}
  \brow{d2}{([yshift=-3.6pt]d1.south west)}{RL and rewards}%
       {Time-R1\,\cite{liu2025timer1}\bul TimeMaster\,\cite{zhang2025timemaster}%
        \bul outcome rewards\,\cite{turtel2025outcome}}
  \brow{d3}{([yshift=-3.6pt]d2.south west)}{Ablation evidence}%
       {no-LLM ablations\,\cite{tan2024actually}%
        \bul in-context limits\,\cite{zhou2025why}}
  \bspine{d1}{d3}
  \colorlet{cX}{cEval}
  \bhdr{h5}{([xshift=-\txnx, yshift=-12pt]d3.south west)}{Evaluation}%
       {\sct{\ref{sec:evaluation}}}{\txcolw}
  \brow{e1}{([xshift=\txnx, yshift=-7.6pt]h5.south west)}{Series benchmarks}%
       {GIFT-Eval\,\cite{aksu2024gifteval}\bul Monash\,\cite{godahewa2021monash}%
        \bul TFRBench\,\cite{ahamed2026tfrbench}\bul FinTSBridge\,\cite{wang2025fintsbridge}}
  \brow{e2}{([yshift=-3.6pt]e1.south west)}{Event benchmarks}%
       {ForecastBench\,\cite{karger2024forecastbench}\bul FutureX\,\cite{zeng2025futurex}%
        \bul MIRAI\,\cite{ye2024mirai}}
  \brow{e3}{([yshift=-3.6pt]e2.south west)}{Calibration \& UQ}%
       {confidence\,\cite{xiong2023can}\bul conformal\,\cite{angelopoulos2021conformal}%
        \bul coverage\,\cite{asch2026rigorous}\bul UQ\,\cite{devic2025from}}
  \brow{e4}{([yshift=-3.6pt]e3.south west)}{Leakage \& cost}%
       {unreported cost\,\cite{moghadasi2026what}\bul saturation\,\cite{yang2024critical}%
        \bul hindcast replay\,\cite{ye2026hindcast}}
  \bspine{e1}{e4}
  \colorlet{cX}{cApp}
  \bhdr{h6}{([xshift=-\txnx, yshift=-12pt]e4.south west)}{Applications}%
       {\sct{\ref{sec:applications}}}{\txcolw}
  \brow{f1}{([xshift=\txnx, yshift=-7.6pt]h6.south west)}{Finance}%
       {FinMem\,\cite{yu2024finmem}\bul FinAgent\,\cite{zhang2024finagent}%
        \bul FinCon\,\cite{yu2024fincon}\bul M6\,\cite{makridakis2024m6}}
  \brow{f2}{([yshift=-3.6pt]f1.south west)}{Weather \& response}%
       {GraphCast\,\cite{lam2023graphcast}\bul Pangu\,\cite{bi2022pangu}%
        \bul GenCast\,\cite{price2024gencast}\bul SIREN\,\cite{ni2026siren}}
  \brow{f3}{([yshift=-3.6pt]f2.south west)}{Health \& epidemics}%
       {COVID Hub\,\cite{cramer2022covidhub}\bul PandemicLLM\,\cite{du2024pandemicllm}%
        \bul EPIFNP\,\cite{kamarthi2021when}}
  \bspine{f1}{f3}
  \coordinate (bt) at ([yshift=-15pt]current bounding box.south -| o);
  \colorlet{cX}{cRisk}
  \bhdr{h7}{(bt)}{Risks \& Open Problems}{\sct{\ref{sec:challenges}}}{\txfullw}
  \qrow{g1}{([xshift=\txnx, yshift=-7.6pt]h7.south west)}{Overconfidence}%
       {market losses\,\cite{cheng2026polybench}\bul solar flares\,\cite{camporeale2025verification}%
        \bul perishable scores\,\cite{gilda2026position}}
  \qrow{g2}{([yshift=-3.6pt]g1.south west)}{Tools \& reflexivity}%
       {tool privilege\,\cite{yang2026when}\bul retrieval filtering\,\cite{lee2026not}%
        \bul LLM crowd aggregation\,\cite{schoenegger2024wisdom}}
  \qrow{g3}{([yshift=-3.6pt]g2.south west)}{Open problems}%
       {ablation dispute\,\cite{qiu2026rethinking}\bul human teams\,\cite{ming2026humancapital}%
        \bul theory\,\cite{sarfati2026probing}}
  \bspine{g1}{g3}
\end{tikzpicture}

%% file: sections/02_methodology.tex
\subsection{Scope}

We include systems where an LLM functionally contributes, through reasoning, representation, or control, to a \emph{scored prediction about a target that is not yet observed at decision time} and can be verified against outcomes using explicit scoring rules. Eligible outputs encompass point predictions, predictive distributions, categories, and event probabilities, excluding non-scorable narrative commentary. Purely numerical forecasting models without language components are excluded from the agent category but retained as baselines. By encompassing both prompt-only models and closed-loop agents, this boundary enables us to examine whether increased agency provides measurable predictive benefits.

The criterion admits targets of either form: numerical targets, typically benchmarked against statistical and foundation models~\citep{gruver2023llmtime,jin2024timellm,ansari2024chronos}, and event targets, typically benchmarked against human forecasters and prediction markets~\citep{zou2022autocast,halawi2024approaching,karger2024forecastbench}. Section~\ref{sec:foundations} formalizes both as one task. We exclude trajectory and motion prediction, which follow distinct task
definitions and evaluation protocols~\citep{xu2025trajectory}, retrospective causal inference, which estimates historical effects rather than future outcomes, and general agent benchmarks without scored predictions.

\subsection{Corpus Construction}

We searched major scholarly databases, preprint platforms, and relevant AI and machine-learning venues using combinations of model terms (\emph{large language model}, \emph{LLM}, \emph{foundation model}, and \emph{agent}), task terms (\emph{forecasting}, \emph{time series}, \emph{prediction}, \emph{event prediction}, and \emph{nowcasting}), and mechanism terms (\emph{prompting}, \emph{retrieval}, \emph{tool use}, \emph{multi-agent}, \emph{calibration}, and \emph{conformal}). The corpus is maintained as a living snapshot of this literature.

We supplemented this search with backward citation search for foundational work in classical forecasting~\citep{box2015time,hyndman2021forecasting}, proper scoring rules~\citep{brier1950verification,gneiting2007strictly,gneiting2007probabilistic}, conformal prediction~\citep{shafer2008tutorial}, forecasting competitions~\citep{makridakis2020m4,makridakis2022m5}, and human superforecasting~\citep{tetlock2015superforecasting}. We screened titles and abstracts before full-text review, retaining studies that propose, evaluate, or analyze LLM-based predictive systems or provide relevant benchmarks, datasets, metrics, or negative results. We excluded papers that use LLMs solely for data cleaning or report generation, position papers without empirical evidence or a new framework, and superseded workshop abstracts. For duplicate records, we cite the most complete published version.

Figure~\ref{fig:corpus} and its accompanying data artifact report the current corpus size and its venue-status, publication-year, and application-domain distributions. Works tied to an application domain are assigned to that domain, while general methods, benchmarks, and foundations are classified as Cross-domain. These quantities may change as the living paper is updated. We explicitly note where contested claims rely solely on preprint evidence.

\input{figures/fig_corpus}

Two main limitations affect the reported results. First, a pilot audit of 12 benchmark papers found substantial disclosure gaps in its sample, including omitted agent scaffolds and no inference-cost reporting among the eight agent papers~\citep{moghadasi2026what}. Consequently, we avoid constructing cross-paper leaderboards. Second, retrospective benchmarks introduce contamination risks through pre-training data exposure or lookahead feature leakage~\citep{guan2026leakage} (see Section~\ref{ssec:formulation}). For such benchmarks, we explicitly note whether studies controlled for contamination.

\subsection{A Unified Formulation}
\label{sec:foundations}
\label{ssec:formulation}

Let $\mathcal{I}_t$ denote the \emph{information set} available strictly before decision time $t$, and let $q$ denote the \emph{forecasting query}, which specifies the prediction target, temporal horizon or event resolution rule, and required output. A forecasting task asks the forecaster to report a predictive distribution for an unobserved target $Z$ using the admissible information
$\mathcal{I}_t$ under the task specification $q$. The target is observed only after the forecast is issued.

When the target is real-valued, $q$ specifies the target series and forecast horizon $H$. Given a series $y_{1:T} \in \mathbb{R}^{T \times d}$, covariates $x_{1:T}$, and unstructured context $c_{1:T}$, the task is
\begin{equation}
\hat{p}\bigl(y_{T+1:T+H} \,\big|\, \mathcal{I}_T\bigr),
\qquad
\mathcal{I}_T = \{y_{1:T}, x_{1:T}, c_{1:T}\},
\label{eq:numerical}
\end{equation}
for forecast horizon $H$. A point forecast reports a summary functional of
$\hat{p}$.

When the target is a discrete event, $q$ specifies the event question together with its resolution criterion and resolution date $\tau$. The task is
\begin{equation}
\hat{\pi}
=
\hat{P}\bigl(Z = 1 \,\big|\, \mathcal{I}_t\bigr),
\qquad t < \tau,
\label{eq:event}
\end{equation}
where $Z \in \{0,1\}$ indicates an affirmative resolution. Multi-outcome questions and questions with numerical intervals generalize this expression to a predictive distribution over outcome space $\mathcal{Z}$.

Equations~\eqref{eq:numerical} and~\eqref{eq:event} are two readouts of one underlying forecasting problem: predicting an unobserved target from admissible information. The appropriate evaluation depends on the form of the forecast. Probabilistic forecasts are evaluated using proper scoring rules~\citep{gneiting2007strictly}: event probabilities commonly use the Brier score~\citep{brier1950verification} or logarithmic score, whereas continuous predictive distributions can be evaluated with the continuous ranked probability score (CRPS). Quantile forecasts are commonly evaluated with pinball loss, while point forecasts use error measures such as MAE, RMSE, MASE, or sMAPE. Proper probabilistic scores reward both calibration and
sharpness and discourage unjustified confidence. For probabilistic forecasts, following \citet{gneiting2007probabilistic}, the central objective is to achieve sharp predictive distributions subject to calibration, namely statistical consistency between predictive distributions and observations. However, existing LLM forecasters often exhibit calibration challenges (Section~\ref{ssec:uq}).

A forecast is admissible only if $\hat{p}$ is measurable with respect to $\mathcal{I}_t$. Leakage undermines evaluation through two main modes: \emph{feature leakage}, where covariates contain lookahead information unavailable at decision time~\citep{guan2026leakage}, and \emph{parametric leakage}, where pre-training data includes benchmark outcomes, which is a risk mitigated by live and simulated benchmarks~\citep{karger2024forecastbench,zeng2025futurex,lee2026forecastbench}. 

\subsection{What Makes a Forecaster an Agent}
\label{ssec:agent-def}

We represent LLM-based forecasting systems within a common agentic framework in which a language model contributes, through reasoning, representation, or control, to a verifiable forecast. The framework accommodates both prompt-only forecasters and systems that update their state through multiple actions and
observations~\citep{chen2026reasoning}. Formally, a system is represented as a tuple $(\pi_\theta, \mathcal{A}, \mathcal{T}, \mathcal{M}, g)$ containing a policy, action space, tools, memory, and predictive readout. Starting from
$s_0 = (q, \mathcal{I}_t)$, it repeats
\begin{equation}
a_k \sim \pi_\theta(\cdot \mid s_k), \quad
o_k = \mathrm{exec}(a_k), \quad
s_{k+1} = u(s_k, a_k, o_k),
\end{equation}
until it emits $\hat{p} = g(s_K)$, where $u$ denotes the state-transition operator induced by actions, observations, and memory updates. Actions can include retrieval, tool calls, memory writes, delegation, and termination. The formulation extends ReAct~\citep{yao2023react} with a predictive readout.
Self-critique methods~\citep{madaan2023selfrefine} and verbal
reinforcement~\citep{shinn2023reflexion} instantiate the update rule $u$, while learned tool invocation~\citep{schick2023toolformer} defines access to $\mathcal{T}$.

Agency is graded: when $K = 1$ and
$\mathcal{A} = \{\textsc{emit}\}$, the system reduces to the prompt-only zero-shot forecaster described in Section~\ref{sec:data}. Controller architectures instead use the LLM to coordinate specialized forecasting models, with their outputs incorporated into the agent state before the final predictive readout $g$ (Section~\ref{sec:hybrid}). We therefore treat agency as a graded property, operationalized primarily through the breadth of the action space $\mathcal{A}$ and the extent of observation-conditioned sequential control; $K$ provides a useful but incomplete proxy for interaction depth. 

Along this continuum, classical statistical models~\citep{box2015time,hyndman2021forecasting} and deep sequence architectures~\citep{zhou2021informer,nie2023patchtst} map numerical history via fixed parameters. Foundation models enable zero-shot transfer but lack query-time parameter adaptation~\citep{ansari2024chronos,das2024timesfm,woo2024moirai}, whereas prompt-only systems such as LLMTime adapt only through their context window~\citep{gruver2023llmtime}. Forecasting-specific retrieval adds selected evidence at inference time~\citep{zhang2025timeraf}, and multi-agent systems extend this process through structured deliberation~\citep{das2026nexus,xiao2024tradingagents}. At a highly agentic end of this continuum, open-web deep-research agents
dynamically expand the evidence incorporated into their agent state, requiring live, contamination-resistant evaluation~\citep{zeng2025futurex,li2026findeepforecast}.

In-context learning augments the model context with examples without updating model weights, while retrieval augments the agent state with selected historical segments or admissible external evidence at inference time. Forecasting retrievers must choose which segments to use~\citep{zhou2026semantics,lee2026not,zhang2025timeraf}; tool-using systems separately must decide whether an invocation is necessary for the current model and task~\citep{cheng2026model}.

Observable traces enable process-fidelity evaluation, but coherent traces do not necessarily ensure faithful reasoning, because agents can produce conclusions unsupported by their preceding steps~\citep{wang2026doing}. This discrepancy is particularly difficult to detect in forecasting, where true outcomes are unknown at decision time. TFRBench addresses part of this gap by evaluating agent-synthesized, numerically grounded traces over real time-series datasets~\citep{ahamed2026tfrbench}.

\subsection{Baselines}
\label{ssec:baseline}

\paragraph{Numerical forecasting.}
Classical Box--Jenkins ARIMA models~\citep{box2015time} and state-space models remain strong, low-cost baselines. M4 showed the competitiveness of combinations and hybrid methods~\citep{makridakis2020m4}, while leading M5 solutions relied heavily on boosted trees, cross-learning, covariates, and ensembles~\citep{makridakis2022m5}. M6 further highlighted the distinction between predictive skill and profitable decisions~\citep{makridakis2024m6}. These numerical baselines cannot directly consume qualitative sources such as news, events, or policy announcements without an additional representation or feature-extraction stage.
Transformer variants target long-horizon forecasting via mechanisms like patch tokenization~\citep{nie2023patchtst} and inverted attention~\citep{liu2024itransformer}. However, a tuned single-layer linear model later matched many leading transformers~\citep{zeng2023dlinear}, highlighting the need for strong baselines and evaluation protocols that do not merely reward model capacity, and that lesson applies directly to LLM-based forecasting.
Time-series foundation models pre-train on heterogeneous corpora to support zero-shot forecasting, whereas prompt-only LLMs show fragile transfer that can be sensitive to small input perturbations and less accurate than statistical baselines~\citep{park2025revisiting}. Prominent architectures tokenize time series via value quantization~\citep{ansari2024chronos}, patch-based decoders~\citep{das2024timesfm}, or any-variate multi-frequency attention~\citep{woo2024moirai}, with broader variants exploring MoE routing, alternative objectives, and scaling~\citep{liu2024moiraimoe,shi2025timemoe,goswami2024moment,liu2025sundial}. Standardized suites evaluate these zero-shot capabilities across diverse domains~\citep{aksu2024gifteval}, with Section~\ref{sec:hybrid} examining their core design trade-offs.
Time-series foundation models (e.g., Chronos and TimesFM) are strong general-purpose baselines for numerical forecasting and are far smaller than frontier LLMs. Consequently, studies proposing LLM-based components must demonstrate clear performance gains over these dedicated baselines, yet Section~\ref{ssec:ablation} finds limited empirical evidence of such improvements.

\paragraph{Event and judgmental forecasting.}
Event forecasting requires different baselines because predictions are
typically probabilities over discrete outcomes rather than numerical
trajectories. General forecasting crowds provide an aggregation-based human
reference~\citep{karger2024forecastbench}, while expert superforecasters
constitute a stronger judgmental baseline~\citep{tetlock2015superforecasting,karger2024forecastbench}.
Automated systems such as Autocast provide complementary machine baselines on
forecasting-tournament questions~\citep{zou2022autocast}. LLM-based forecasters may approach crowd-level performance on selected question sets, but expert forecasters remain an important comparator~\citep{halawi2024approaching,karger2024forecastbench}. Market forecasts provide an additional reference for market-settled questions, although forecast quality should be evaluated separately from downstream returns.
Section~\ref{sec:evaluation} examines these comparisons and their associated evaluation protocols.

%% file: figures/fig_corpus.tex
\begin{figure}[t]
\centering
\includegraphics[width=\textwidth]{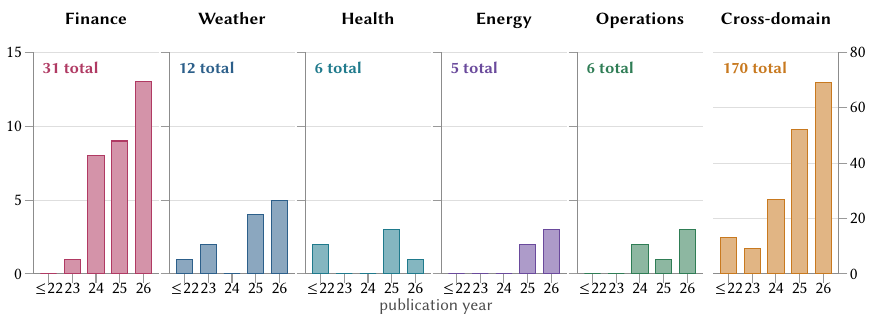}
\caption{Publication-year distribution of references grouped by application
domain. The panels cover Finance, Weather, Health, Energy,
Operations, and Cross-domain. The applied panels share one vertical scale,
while Cross-domain uses a larger scale. 
}
\label{fig:corpus}
\end{figure}

%% file: sections/04_data-prompting.tex
We categorize forecasting architectures by how components interface with the
LLM: (1) standalone LLMs using specific representations and prompts; (2)
tool-augmented systems integrating programs, retrieval, or live evidence into
the agent state; and (3) hybrid models pairing the LLM with statistical, deep
learning, or time-series foundation backbones. These categories capture
dominant coupling patterns rather than mutually exclusive system classes,
since a system may combine retrieval, multi-agent orchestration, structured
knowledge, and numerical forecasting backbones. Each setup adds distinct
evaluation constraints: retrieval must prevent lookahead leakage, and hybrid
designs must demonstrate gains over their standalone numerical baselines.
These architectures realize the adaptive forecasting framework of
\citet{cheng2026agentic}. Throughout this section, we focus primarily on
inference-time architecture and component coupling. Training-time parameter
adaptation is discussed systematically in Section~\ref{sec:training}, although
some systems reviewed here include learned or fine-tuned interfaces.

\subsection{Serialization, Tokenization, and Prompt Content}
\label{sec:data}
\label{ssec:tokenisation}

Because text transformers lack native representations for real-valued quantities, their performance hinges on tokenization and serialization choices. The upper block of Table~\ref{tab:architectures} categorizes these designs across three axes: numerical formatting, token segmentation, and accompanying contextual text.

\begin{table}[t]
\centering
\scriptsize
\setlength{\tabcolsep}{2.2pt}
\renewcommand{\arraystretch}{0.93}
\caption{Representative forecasting systems organized by representation and control flow. The upper block varies forecast representation and context, whereas the lower block varies control flow from fixed refinement to open-web search. $K$ counts model calls (Section~\ref{ssec:agent-def}).}
\label{tab:architectures}
\begin{tabularx}{\textwidth}{@{}>{\hsize=1.05\hsize\raggedright\arraybackslash}XW{0.80}W{1.10}W{1.30}W{0.75}@{}}
\toprule
System & Family & Forecast input / evidence & Forecast / control strategy & Calls / loop $K$ \\
\midrule
LLMTime~\citep{gruver2023llmtime} & Single-agent & Space-separated digit series & Zero-shot sequence continuation & 1 \\
TOKON~\citep{yang2025tokon} & Serialization & One token per series value & Vocabulary-fitted normalization & 1 \\
Patch prompting~\citep{bumb2025forecasting} & Serialization & Patched nearest-neighbor series & Trend + residual components & 1 \\
PromptCast~\citep{xue2023promptcast} & Single-agent & Digits + task sentence & Fixed question template & 1 \\
Prompt mining~\citep{xue2024prompt} & Single-agent & Numerical text & Template search & 1 \\
Noise injection~\citep{yin2026enhancing} & Prompt perturbation & Perturbed numerical text; synthetic noise & Sequence continuation & 1 \\
LSTPrompt~\citep{liu2024lstprompt} & Single-agent & Numerical text + horizon & Horizon-specific prompts & 1 \\
MAP4TS~\citep{lee2025map} & Prompt design & Numerical text + dataset context & Global, local + statistical views & 1 \\
STELLA~\citep{fan2025stella} & Single-agent & Series + component summaries & Mined semantic abstractions & 1 \\
LLM-Mixer~\citep{kowsher2024llm} & Learned encoder & Multiscale embeddings + descriptors & Task-specific text prompt & 1 \\
UniCast~\citep{park2025unicast} & Learned prompt & Series + image + text embeddings & Instance-conditioned soft prompt & 1 \\
MORFI~\citep{khezresmaeilzadeh2025morfi} & Single-agent & Chart + price text & Multimodal chain-of-thought & 1 \\
LLM--transformer fusion~\citep{su2025fusing} & Fused encoder & Projected patches + Transformer features & Gated semantic / temporal fusion & 1 \\
\midrule
FLAIRR-TS~\citep{jalori2025flairr} & Forecaster + refiner & Numerical series & Test-time prompt optimization & Fixed, small \\
CastFSR~\citep{tao2026castfsr} & Single-agent loop & Series + context & Fast / slow / reflective modes & Adaptive \\
Cross-RAG~\citep{lee2026not} & Retrieval-augmented & Numerical series & Query--retrieval cross-attention & 1 \\
SERAF~\citep{zhou2026semantics} & Retrieval-augmented & Series + text & Joint numerical + text retrieval & 1 \\
TimeRAF~\citep{zhang2025timeraf} & Retrieval-augmented & Numerical series & Learnable retriever & 1 \\
Nexus~\citep{das2026nexus} & Multi-agent & Series + event text & Macro / micro role decomposition & Role-dependent \\
TradingAgents~\citep{xiao2024tradingagents} & Multi-agent & Market + text & Role-based deliberation & Role-dependent \\
ElliottAgents~\citep{chudziak2025elliottagents} & Multi-agent & Market + text & Natural-language debate & Role-dependent \\
TimeSeriesScientist~\citep{zhao2025tsscientist} & Multi-agent pipeline & Series + diagnostics & Curate, plan, forecast + report & Four staged roles \\
CastFlow~\citep{pan2026castflow} & Role-specialized workflow & Series + text & Plan, act, forecast + reflect & Fixed reflective cycle \\
KairosAgent~\citep{feng2026kairosagent} & Reasoner + forecaster & Series + text & Fused semantic reasoning & Multi-turn; per-turn credit \\
TimeClaw~\citep{li2026timeclaw} & Generalist runtime & Numerical arrays + text & Executable temporal tools & Variable; tool-bounded \\
SEA-TS~\citep{xu2026seats} & Code-writing agent & Numerical series + generated code & Evolutionary algorithm search & Unbounded search \\
AIA Forecaster~\citep{alur2025aiaforecaster} & Multi-agent research & Live web news & Search + reconciliation & Unbounded; arbitrated \\
ForecastAgentSearch~\citep{cai2026forecastagentsearch} & Proposed multi-expert design & Expert-agent pool & Proposed expert ranking & Not evaluated \\
Argumentative MAS~\citep{gorur2025retrieval} & Multi-agent judgmental & Retrieved news & Structured argument & Debate rounds \\
ForecastCompass~\citep{chang2026forecastcompass} & Memory-augmented research & Question context + memory & Factor + calibration recall & Memory-fed \\
\bottomrule
\end{tabularx}
\end{table}
PromptCast and LLMTime represent two contrasting strategies within this
design space. PromptCast frames forecasting as question-answering via sentence templates~\citep{xue2023promptcast}, whereas LLMTime feeds a frozen LLM bare digit sequences with space-delimited digits to control tokenization~\citep{gruver2023llmtime}. 
The contrast matters because sentence framing exposes task semantics, while digit spacing produces more regular token boundaries at greater context cost; neither choice establishes an advantage over a strong numerical baseline.

\vspace{-0.1in} \paragraph{Numerical Encoding}
Alternative tokenizations replace natural language with specialized units. Chronos quantizes values into learned vocabularies via backbone retraining~\citep{ansari2024chronos}, serving as a key baseline (Section~\ref{ssec:baseline}). Patch tokenization, adapted from sequence models~\citep{nie2023patchtst}, groups time windows to provide a more efficient representation of local temporal patterns~\citep{bumb2025forecasting}, while vision--language approaches process rendered charts to infer trends directly from pixels~\citep{khezresmaeilzadeh2025morfi}. Surrounding text prompts also heavily impact performance; because manual templates are often suboptimal, some comparisons may reflect prompt tuning rather than fundamental model gains~\citep{xue2024prompt}. Finally, irregular formats pose distinct challenges, such as asynchronous event streams lacking fixed-width serializations~\citep{gupta2025last} or clinical records arriving as temporally annotated narrative text~\citep{noroozizadeh2025temporally}.

General-purpose language tokenizers are optimized for natural-language statistics rather than numerical structure, which can lead to inconsistent tokenization of adjacent values, loss of numerical locality, and inefficient token usage. LLMTime inserts spaces to obtain per-digit tokenization, improving positional regularity but adding nuisance tokens and lengthening the sequence~\citep{gruver2023llmtime}. Alternatively, TOKON normalizes each value into a single token, reducing sequence length by $2\times$ to $3\times$ while improving accuracy~\citep{yang2025tokon}, which highlights key trade-offs between context efficiency and numerical granularity.

Clean token boundaries do not ensure that embeddings preserve numerical order or distance. One hybrid design applies reversible normalization, sends projected numerical patches to an LLM branch, and fuses them with a temporal transformer~\citep{su2025fusing}; related hybrids likewise preserve a dedicated numerical pathway rather than relying on text serialization alone~\citep{xiong2025beyond} (Section~\ref{sec:hybrid}).

Because quadratic self-attention costs force a trade-off between numerical precision and lookback context length, patching groups adjacent observations into tokens to reduce attention cost and expand the effective lookback window~\citep{nie2023patchtst}, while encoding granularity can limit the resolution of predictive distributions~\citep{gruver2023llmtime} (Section~\ref{ssec:uq}). A key evaluation gap remains: existing studies often change serialization together with prompts and backbones rather than isolating encoding effects on a fixed benchmark.
This coupling makes forecast error an incomplete diagnostic: a model may fail because it misreads the encoded history, because it cannot extrapolate the pattern, or both. Fixed-backbone comparisons should therefore pair forecasting with an input-reconstruction task, separating numerical fidelity from predictive ability.

\vspace{-0.1in} \paragraph{Prompt and Context Design}
Beyond value encoding, methods leverage remaining context space to supply structured task semantics. MAP4TS incorporates dataset-level context, local dynamics, statistical properties, and temporal dependencies into its prompt~\citep{lee2025map}, while STELLA extracts trend, seasonal, and residual components as structured semantic summaries~\citep{fan2025stella}. Both frameworks perform signal decomposition externally, using the extracted features to guide LLM predictions.

LSTPrompt separates prompts by forecast horizon and periodically prompts the model to reassess its strategy~\citep{liu2024lstprompt}, while LLM-Mixer multi-scales temporal features before conditioning the backbone with task-specific text prompts~\citep{kowsher2024llm}. Both adapt chain-of-thought principles~\citep{wei2022cot} to task decomposition. Unlike standard reasoning tasks, however, forecasting intermediate steps cannot be validated before target observation, which is a limitation that multi-turn control-flow architectures seek to address (Section~\ref{ssec:refinement}).

UniCast uses a learned distiller to generate instance-conditioned prompts from time-series, image, and text inputs~\citep{park2025unicast}, bridging manual prompting and parameter adaptation. Related approaches partially fine-tune normalization layers~\citep{zhou2023onefitsall} or align time-series patches with textual prototypes~\citep{jin2024timellm}. Because these methods incur training overheads absent in zero-shot prompting, Section~\ref{sec:training} classifies them under parameter adaptation.

\vspace{-0.1in} \paragraph{Robustness to Perturbations}
Reported input noise effects remain inconsistent: injected noise improved zero-shot accuracy in one study~\citep{yin2026enhancing}, while small perturbations sharply degraded forecasts in another~\citep{park2025revisiting}. This discrepancy likely stems from interactions between perturbations and value encodings, mirroring broader degradation in LLM reasoning under noisy inputs~\citep{tian2025large} and highlighting unresolved sensitivity to surface-level changes.

On standard numerical benchmarks, removing or replacing the LLM component can leave accuracy unchanged or improved~\citep{tan2024actually}, indicating that representation studies may optimize an interface to a non-contributing module. This occurs either because purely numerical benchmarks lack load-bearing text or because representation is not the limiting factor. Isolating these mechanisms requires a factorial design varying encoding quality and LLM presence on text-rich tasks, which is an experiment built on existing components~\citep{tan2024actually,yang2025tokon} but not yet reported (Section~\ref{sec:benchmarks}).

\subsection{Single Pass, Refinement, and Multi-Agent Orchestration}
\label{ssec:refinement}
\label{ssec:multi-agent}

Representation and control flow constrain one another. A loop cannot recover information discarded during serialization, while accurate encoding alone may provide limited benefit if the architecture cannot revise an incorrect initial forecast. We distinguish single-pass calls, bounded refinement, role-specialized multi-agent workflows, and open-web research loops. Their call count $K$ is not a complete cost measure because communication topology and retrieved context also determine latency and token use~\citep{orogat2026understanding}. Open-web loops additionally vary the evidence set across queries, which improves access to post-training events but weakens reproducibility.

\vspace{-0.1in} \paragraph{Refinement and Adaptive Routing}
Iterative generation--evaluation loops ($K>1$) allow test-time adaptation without gradient updates. FLAIRR-TS reports benchmark improvements from retrieval-augmented refinement, but each pass adds model calls and its cost-effectiveness across backbones and datasets remains unresolved~\citep{jalori2025flairr}.
Dynamic routing (e.g., CastFSR~\citep{tao2026castfsr}) allocates compute by switching between fast, contextual, and reflective modes based on series complexity. While promising for cost efficiency (Section~\ref{sec:metrics}), aggregate benchmarks have not yet established router reliability on challenging cases.
TimeRAF shows that selectively weighting retrieved historical segments can improve forecasts~\citep{zhang2025timeraf}. TFRBench separately evaluates grounded trace quality and trace-conditioned forecast gains, but does not isolate retrieval as the causal mechanism~\citep{ahamed2026tfrbench} (Section~\ref{ssec:tool-augmented}).

\vspace{-0.1in} \paragraph{Role Specialization and Native Execution}
Multi-agent frameworks distribute compute across specialized roles, either by structural signal decomposition (e.g., macro/micro dynamics in Nexus~\citep{das2026nexus}) or domain-role division (e.g., analysts and traders in TradingAgents~\citep{xiao2024tradingagents}). Because these setups encode distinct domain assumptions, empirical validation of one role architecture does not generalize to others.
CastFlow assigns planning, action, forecasting, and reflection to a workflow in which a dedicated forecaster retains responsibility for numerical prediction~\citep{pan2026castflow}. KairosAgent similarly couples a language reasoner to a time-series foundation model but uses turn-level credit to distinguish useful reasoning steps from the final trajectory~\citep{feng2026kairosagent}. This division of labor is motivated by weak LLM ablation results: language modules can process context and control, while numerical modules
preserve temporal inductive biases. It also creates a new burden, since final-error metrics alone cannot determine whether a failure arose from reasoning, handoff, or prediction.
TimeClaw identifies datatype and process misalignment, together with temporal-resolution distortion, when general text interfaces mediate time-series work~\citep{li2026timeclaw}. On TimeSage-MT, models are relatively stronger at memory and chaining but remain weak in numerical accuracy and analytical grounding~\citep{kong2026timesagemt}. These results motivate evaluating native execution and reasoning quality separately from final prediction accuracy.

\subsection{Tool Use, Program Synthesis, and Open-Web Research}
\label{ssec:tool-augmented}

An LLM moves beyond prompt-only forecasting when it calls APIs, executes code, or queries search engines to expand its accessible information state during inference (Table~\ref{tab:hybrid}). These systems build on foundational agentic mechanisms such as tool integration~\citep{schick2023toolformer}, interleaved reasoning--action loops~\citep{yao2023react}, self-critique/reflection~\citep{madaan2023selfrefine,shinn2023reflexion}, and multi-agent orchestration~\citep{wu2023autogen}.

\begin{table}[t]
\centering
\scriptsize
\setlength{\tabcolsep}{2.4pt}
\renewcommand{\arraystretch}{0.93}
\caption{LLM couplings: external evidence (top), numerical foundation-model
baselines (middle), and hybrid or controller forecasters (bottom). Final-block
systems require comparisons with the middle block, not only classical baselines.}
\label{tab:hybrid}
\begin{tabularx}{\textwidth}{@{}>{\hsize=1.00\hsize\raggedright\arraybackslash}XW{0.75}W{1.10}W{1.00}W{1.15}@{}}
\toprule
System / study & Family & Coupled component & LLM role & Forecast target / setting \\
\midrule
TimeRAF~\citep{zhang2025timeraf} & Retrieval & Knowledge base + foundation model & Retrieval for the base model & Zero-shot series; retrieved evidence \\
Cross-RAG~\citep{lee2026not} & Retrieval & Retrieved series & Query--retrieval cross-attention & Time-series benchmarks \\
\citet{zhou2024synergizing} & Structured knowledge & LBSN knowledge graph & Meta-path graph reasoning & Socioeconomic indicators; benchmarks \\
Nexus~\citep{das2026nexus} & Structured knowledge & News + event streams & Staged multi-agent decomposition & Time-series benchmarks \\
EventCast~\citep{hu2026eventcast} & Structured knowledge & Demand model + event text & Extracts event evidence & E-commerce demand; case study \\
\midrule
Chronos-2~\citep{ansari2025chronos2} & Foundation model & Group attention; T5 lineage & None; general baseline & Cross-domain zero-shot; covariates \\
Toto 2.0~\citep{khwaja2026toto2} & Foundation model & Decoder-only; 4M--2.5B parameters & None; scaling baseline & Telemetry zero-shot; observability \\
Timer-S1~\citep{liu2026timers1} & Foundation model & 8.3B sparse mixture of experts & None; frontier baseline & GIFT-Eval state of the art; univariate \\
Xihe~\citep{sun2025xihe} & Foundation model & Hierarchical interleaved attention & None; scaling baseline & Size-scaling zero-shot; univariate \\
Reverso~\citep{fu2026reverso} & Foundation model & Efficient 2.6M-parameter backbone & None; low-cost baseline & Compute-limited zero-shot; univariate \\
CoRA~\citep{qin2025cora} & Foundation-model adapter & Frozen TSFM + modality encoders & GCE + zero-initialized injection & Covariate-aware forecasting \\
TimesFM~\citep{das2024timesfm} & Foundation model & Patched decoder-only transformer & None; purpose-built baseline & Cross-domain zero-shot; univariate \\
Moirai~\citep{woo2024moirai} & Foundation model & Any-variate attention encoder & None; purpose-built baseline & Cross-frequency zero-shot; multivariate \\
\midrule
UniCast~\citep{park2025unicast} & Hybrid & Frozen foundation model & Instance-conditioned context & General forecasting; series + text + images \\
LLIAM~\citep{germanmorales2024transfer} & Hybrid & Foundation-model backbone & LoRA-adapted forecaster & General forecasting; text-serialized series \\
\citet{benjamim2026soybean} & Hybrid & LSTM & Sentiment feature extractor & Soybean futures; news text \\
\citet{niu2026forecasting} & Controller & Prophet, XGBoost, LSTM pool & Coordinator over baseline forecasts & Low-carbon microgrids; synthetic data \\
\bottomrule
\end{tabularx}
\end{table}

\vspace{-0.1in} \paragraph{Feedback and Tool Choice}
Unlike many tool-use tasks where within-episode feedback is immediate, prospective forecasting targets may only resolve after deployment. FLAIRR-TS instead uses retrieval and validation-driven iterative refinement and reports lower benchmark MAE, at the cost of repeated model calls~\citep{jalori2025flairr}. The outcome delay also motivates consistency objectives that reward alignment between reasoning and action before resolution~\citep{chen2026reasoning}.
Because correctness is unavailable within the episode, validation can reject malformed outputs or inconsistent rationales but cannot verify the prediction itself.

Tool value depends on the model and task: stronger backbones may bypass tools that weaker models require, so fixed budgets do not transfer across architectures~\citep{cheng2026model}. Agents also select over-privileged tools when cheaper options suffice~\citep{yang2026when}, adding safety and inference costs. Policies should therefore govern both tool availability and call
necessity.

\vspace{-0.1in} \paragraph{Retrieval and Program Synthesis}
TimeRAF retrieves historical segments, Cross-RAG downweights unhelpful series, and joint shape-text retrieval addresses nonstationarity errors~\citep{zhang2025timeraf,lee2026not,zhou2026semantics}. Evidence must also satisfy decision-time constraints~\citep{guan2026leakage}. The Bayesian Linguistic Forecaster limits context growth by updating a compressed
probability-and-text belief state~\citep{murphy2026agentic}; its long-horizon trade-off against raw retrieval remains untested.
Relevance without timestamp admissibility can raise apparent accuracy by importing unavailable evidence, so retrieval quality and temporal validity require separate ablations.

Code-writing agents can generate executable forecasting programs rather than emitting direct predictions. For example, SEA-TS searches over algorithmic candidates, evaluates them on held-out data, and evolves code based on error metrics~\citep{xu2026seats}. The resulting program is inspectable, executable, reusable without recurring frontier-model calls, and completely bypasses the numerical tokenization pitfalls detailed in Section~\ref{sec:data}. Held-out error supplies immediate feedback that direct event forecasting lacks, but repeated search against one validation split can overfit model selection. Program inspection improves auditability without by itself guaranteeing temporal validity or out-of-distribution performance.

\vspace{-0.1in} \paragraph{Open-Web Research and Memory}
News-search systems can approach human crowds on some event-question sets~\citep{halawi2024approaching,alur2025aiaforecaster}. Implementations use supervisor reconciliation or structured argumentation~\citep{alur2025aiaforecaster,gorur2025retrieval}; ForecastAgentSearch proposes expert-agent ranking without forecasting experiments~\citep{cai2026forecastagentsearch}. Live, future-resolving questions reduce the risk of parametric leakage~\citep{zeng2025futurex,li2026findeepforecast}, but open-web
access adds security risk (Section~\ref{ssec:tool-risk}).
ForecastCompass stores predictive factors separately from calibration lessons and updates both after question resolution, providing a non-gradient form of forecasting-specific memory~\citep{chang2026forecastcompass}. Whether such memory tracks distribution shifts better than fixed model weights remains untested.

\subsection{Structured Knowledge and Live Evidence}

Tool libraries grant agents operational actions, whereas knowledge graphs provide relational evidence to interpret. One socioeconomic prediction system uses an LBSN graph with dynamic subgraph and meta-path reasoning~\citep{zhou2024synergizing}. Separately, \citet{li2026grounding} prove an evidence-relative trade-off for incomplete knowledge graphs: a deterministic grounding rule cannot reject every unsupported trajectory while retaining every valid trajectory whose evidence is missing. Applying that result to scored forecasting remains an open problem.

\vspace{-0.1in} \paragraph{Knowledge Validity}
\citet{theologitis2025thucy} translate natural-language claims into SQL queries and return exact supporting or counterexample records from relational databases. Adapting this verification pattern to filter inconsistent candidate forecasts, rather than generating predictions directly from incomplete evidence, remains largely unexplored.
EventCast draws future operational events from an existing expert database and converts them into summaries for a demand model~\citep{hu2026eventcast}. At a higher level, \citet{nitu2026multimodal} report attention-based fusion of structured ERP sales records with LLM-derived embeddings from external text. Neither result establishes query-level guarantees for source coverage or temporal validity.

\vspace{-0.1in} \paragraph{Streaming and Live Evidence}
Time-aligned news can provide additional predictive signals beyond numerical energy-price history, although current evidence is retrospective rather than a deployed stream~\citep{bi2025nsw}. Architectures incorporate event context through staged components~\citep{das2026nexus} or feed extracted news signals to numerical forecasting models~\citep{hussain2026improving,xiong2025beyond}. A renewable-energy review identifies edge, IoT, latency, and interoperability constraints as deployment challenges rather than evaluated routing gains~\citep{manjunath2026llm}.
Live protocols reduce the risk of leakage through unresolved questions, daily ingestion,
timestamp-locked snapshots, live market states, or generated
tasks~\citep{karger2024forecastbench,zeng2025futurex,cheng2026polybench,
yu2025livetradebench,li2026findeepforecast}. Foresight Arena's prospective
on-chain results remain pending~\citep{nechepurenko2026foresight}, while
retrospective masking retains residual risk~\citep{ma2026oracleproto}. FutureX
also documents fabricated or stale web evidence~\citep{zeng2025futurex};
vetted corpora trade coverage for safety.
Explicit propagation of source validity or evidence decay into predictive distributions remains uncommon among current agents. Robust retrieval therefore requires provenance, timestamps, and expiration criteria.

\subsection{Hybrid Pairing with Statistical and Foundation Models}
\label{sec:hybrid}
\label{ssec:tsfm}

Hybrid systems pair LLMs with quantitative models for feature extraction, control, or end-to-end forecasting. They should be evaluated against strong numerical controls such as Chronos-2 and TimesFM~\citep{ansari2025chronos2,das2024timesfm}. M4 and M5 show the competitiveness of combinations, covariates, trees, and ensembles~\citep{makridakis2020m4,makridakis2022m5}; downstream decision results must also be separated from forecast error~\citep{zhang2023hybrid}. Lower point error need not improve operational utility, and an expensive LLM may only reproduce covariates already available to the numerical model.

One narrow and testable role is text-derived feature extraction. An LLM can convert news into sentiment for an LSTM, or convert promotional calendars into structured events for a demand model~\citep{benjamim2026soybean,hu2026eventcast}. This arrangement reuses one language-model call across many horizons and retains the numerical model's inductive biases. A key limitation is compression: a scalar sentiment or event label can discard timing, uncertainty, and interactions present in the source text. The informative ablation therefore compares the text-derived feature with strong numerical covariates, not only with a context-free model.

\vspace{-0.1in} \paragraph{Numerical Backbones and Controllers}
\citet{mahmud2025hybrid} compare ARIMA, LSTM, and their hybrid, while \citet{qureshi2024statistical} compare ARIMA, XGBoost, and a hybrid. LLM-assisted multi-backbone systems instead use language context alongside numerical priors or candidate models~\citep{niu2026forecasting}. M4 and M5 provide shared data and metrics but do not causally isolate each pipeline component~\citep{makridakis2020m4,makridakis2022m5}; most LLM-augmented hybrids likewise compare only against bare numerical models. The marginal value of language therefore remains confounded by text content, parameter scale, and temporal generalization.
\citet{asaad2026regime} condition a frozen LLM on estimated market regimes and in-context examples to predict next-day realized variance directly; the LLM does not control a separate numerical forecaster. The reported LLM predictor outperforms classical baselines, while a separate quantitative study estimates latent regimes from return dynamics~\citep{tian2026regime}. These results do not isolate whether the LLM gain comes from regime labels, demonstration matching, or language pre-training.
\citet{liao2026bridging} use an LLM to revise air-ticket sales forecasts from business context after a numerical model produces its initial prediction. This contextual revision is distinct from classical statistical--neural residual modeling, and its retrospective case studies do not establish production deployment or general gains at matched cost. Evaluating a controller against only a fixed default model conflates routing quality with candidate-pool performance; proper evaluation requires comparisons with both random selection and an oracle selector.

\vspace{-0.1in} \paragraph{Foundation-Model Baselines}
Foundation models use distinct representations and heads. Chronos quantizes values, TimesFM maps continuous patches to point forecasts, and Sundial generates raw continuous values~\citep{ansari2024chronos,das2024timesfm,liu2025sundial}. Timer uses autoregressive generation, scaled in Timer-S1 with sparse MoE routing~\citep{liu2024timer,liu2026timers1}; alternatives include
MOMENT's masked reconstruction, Moirai's any-variate attention, and other MoE designs~\citep{goswami2024moment,woo2024moirai,shi2025timemoe}. Frozen feature quality also correlates across forecasting and classification tasks~\citep{auer2025pre}. Direct distributional heads differ from post-hoc intervals (Section~\ref{ssec:formulation}). Within-family evaluations of Toto 2.0, Xihe, and Timer-S1 report improvements with increasing scale, but these curves do not establish the best cross-architecture accuracy at a fixed budget.

\vspace{-0.1in} \paragraph{Cross-Modal Interfaces}
LLIAM uses low-rank adaptation to fine-tune backbones (Section~\ref{sec:training})~\citep{germanmorales2024transfer}. Representation-alignment methods address modality mismatch~\citep{ye2024ptlm}, whereas gated fusion combines LLM semantic representations with temporal-Transformer features~\citep{su2025fusing}. TimeRAF offers nonparametric retrieval of historical series without backbone updates~\citep{zhang2025timeraf}. UniCast is not a retrieval method: it learns an instance-conditioned prompt and modality router around a frozen time-series foundation model~\citep{park2025unicast}.

%% file: sections/08_training.tex
Section~\ref{sec:architectures} organized forecasting systems primarily by their inference-time architecture and component coupling. This section instead focuses on how model parameters, interfaces, and policies are adapted during training, and how these choices affect forecasting behavior and inference cost. The supervision structure also dictates optimization dynamics: step-wise numerical targets provide dense rewards, whereas binary events are resolved only after long delays, altering credit assignment and calibration dynamics.

\subsection{Adaptation and Inference-Time Compute}

Most adaptation methods freeze the pretrained LLM weights and train a lightweight interface.
GPT4TS updates normalization and positional-embedding parameters, whereas Time-LLM freezes the LLM weights and trains a reprogramming layer that maps input patches into the LLM embedding space, together with task-specific prompt context. Thus, GPT4TS adapts selected parameters inside the pretrained LLM, while Time-LLM adapts the input interface around a frozen LLM (Table~\ref{tab:training}).
This distinction affects transfer: updating selected model parameters may fit target-specific distribution shifts but can reduce the reusability of pretrained representations, whereas input reprogramming leaves the LLM weights unchanged but remains constrained by the pretrained embedding space, tokenization, and context length.

\begin{table}[t]
\centering
\scriptsize
\setlength{\tabcolsep}{2.3pt}
\renewcommand{\arraystretch}{0.93}
\caption{Forecasting adaptation from lightweight interfaces to reinforcement
learning on delayed events. Numerical horizons yield dense loss; binary events
yield one outcome at resolution.}
\label{tab:training}
\begin{tabularx}{\textwidth}{@{}>{\hsize=0.90\hsize\raggedright\arraybackslash}XW{1.00}W{1.00}W{1.15}W{0.95}@{}}
\toprule
Method & Adaptation & Updated component & Training / calibration signal & Connected modalities \\
\midrule
GPT4TS~\citep{zhou2023onefitsall} & Partial fine-tuning & Normalization + positional layers & Forecast loss (dense) & Values to token embeddings \\
Time-LLM~\citep{jin2024timellm} & Reprogramming & Patch reprogramming layer & Forecast loss (dense) & Patch to text prototype \\
TEST~\citep{sun2024test} & Embedding alignment & Series encoder & Contrastive + forecast losses & Series to text prototype \\
TEMPO~\citep{cao2024tempo} & Prompt pool + fine-tuning & Decomposition prompts & Forecast loss (dense) & Trend / seasonality to prompt \\
CALF~\citep{liu2024calf} & Cross-modal fine-tuning & Dual-branch adapters & Feature + output alignment & Numerical to text branch \\
AutoTimes~\citep{liu2024autotimes} & Autoregressive adaptation & Projection layers & Next-token forecast loss (dense) & Series as token sequence \\
LLIAM~\citep{germanmorales2024transfer} & Low-rank adaptation & LoRA matrices & Forecast loss (dense) & Text-serialized series \\
Time-R1~\citep{liu2025timer1} & Reinforcement learning & Full policy & Temporal rule reward (dense) & Reasoning to timestamped fact \\
TimeMaster~\citep{zhang2025timemaster} & SFT then RL & Full policy & Format + accuracy reward (dense) & Series image to reasoning \\
TimeRFT~\citep{li2026timerft} & Reinforcement fine-tuning & TSFM policy & Per-step / variable reward & Numerical error to reward \\
CastFlow~\citep{pan2026castflow} & SFT then workflow RL & Role-specialized policies & Workflow reward (sparse) & Context to role composition \\
Outcome-RL~\citep{turtel2025outcome} & GRPO + ReMax & Full policy & Delayed Brier reward (sparse) & News evidence to probability \\
Question synthesis~\citep{chandak2025scaling} & GRPO at scale & Full policy & Accuracy + Brier reward (sparse) & Retrieved news to probability \\
FutureWorld~\citep{han2026futureworld} & Live RL environment & Full policy & Delayed outcome reward (sparse) & In-loop retrieval to probability \\
Market reward~\citep{levy2026eventforecasting} & Reward substitution & Full policy & Market-price reward (dense) & Market state to probability \\
RLCR~\citep{damani2025beyond} & RL with confidence target & Policy output sequence & Correctness + proper-score reward & Reasoning to stated confidence \\
Calibrated VR~\citep{singh2026verifiable} & Masked verifiable-reward RL & Gradient-masked policy & State-conditioned empirical-rate reward & Evidence state to probability \\
Beta--Bernoulli~\citep{dai2026betabernoulli} & Post-hoc correction & External calibrator & Outcomes + crowd forecasts & Score to calibrated probability \\
\bottomrule
\end{tabularx}
\end{table}

\vspace{-0.1in} \paragraph{Learned Interfaces}
Numerical and token spaces can be aligned by contrastive or branch objectives
~\citep{sun2024test,liu2024calf}, prompt embeddings
~\citep{cao2024tempo,pan2024s2ipllm}, or direct autoregressive tokenization
~\citep{liu2024autotimes}.
These interfaces should be compared using the same pretrained LLM, data splits, preprocessing, and tuning budget, rather than only against independently tuned published baselines. Otherwise, an apparent gain may reflect differences in model size, pretraining data, preprocessing, or tuning budget rather than improved cross-modal alignment.
LoRA reduces trainable parameters through low-rank adaptation or
importance-based module selection~\citep{germanmorales2024transfer,li2025trace}.
Updating more parameters may better fit target-specific distribution shifts, but can also reduce the transferability of pretrained representations~\citep{auer2025pre,ye2024ptlm}.
CoRA trains causal condition injection over frozen encoders, while MM-PostTrain
trains an LLM reviser over a numerical prior
~\citep{qin2025cora,liu2026mmposttrain}. ChatTS uses synthetic time series with known attributes to train cross-modal alignment, whereas STRIDE distills reasoning traces into continuous encoder representations to reduce inference-time chain-of-thought overhead~\citep{xie2025chatts,ahamed2026stride}. Their generalization under shift
remains unverified~\citep{xie2026posttraining}.

\vspace{-0.1in} \paragraph{Inference-Time Adaptation}
Inference-time compute yields mixed results: verified reasoning traces can improve performance, whereas excessive reasoning tokens may reduce accuracy, extended deliberation may worsen calibration, and tested prompt interventions often fail to improve accuracy~\citep{ahamed2026tfrbench,zhou2026overthinking,nel2025kalshibench,schoenegger2025prompteng}. Repeated sampling offers little numerical diversity,
whereas input perturbation and contextual gating can improve selected
forecasts~\citep{hua2026diversified,pandey2026gitco}. Useful compute therefore
diversifies or filters information rather than repeating unchanged inputs.
Forecasting systems can also externalize experience as reusable memory instead of updating model weights~\citep{zhou2026externalization}. Residual replay buffers provide another form of online adaptation while leaving the forecasting model weights unchanged~\citep{dai2026orca}. The relative robustness of these methods to abrupt regime shifts and gradual novelty has not yet been evaluated.

\subsection{Reinforcement Learning and Calibration}
\label{ssec:rl-outcome}
\label{ssec:train-calibration}

Supervised fine-tuning optimizes forecast outputs from labeled targets, whereas reinforcement learning optimizes the expected reward of sampled trajectories and may affect both intermediate actions and final outputs. For online binary-event forecasting, outcome rewards are sparse and delayed. A forecast of 0.7 is calibrated only in aggregate; under perfect calibration, approximately 30\% of such events resolve negatively. Proper
scores such as negative Brier, optimized through GRPO or ReMax, reward
probability estimation rather than raw confidence
~\citep{gneiting2007strictly,turtel2025outcome}; one 14B model matches the reported accuracy of larger frontier models while achieving better calibration. This result suggests, but does not establish, that reward design can partly offset model-scale differences.

\vspace{-0.1in} \paragraph{Outcome and Market Rewards}
Resolved events provide automated ground-truth labels over time~\citep{turtel2026futureaslabel}. Synthesizing tens of thousands of news-based forecasting questions enables RL with accuracy and Brier rewards; the resulting model is evaluated on FutureX and out-of-distribution calibration, but live-capital and expert-tournament transfer remain untested~\citep{chandak2025scaling}.
FutureWorld lets a policy retrieve evidence for open questions and receive an outcome reward after resolution~\citep{han2026futureworld}. Outcome-only rewards are sparse and high-variance and provide weak credit assignment for intermediate actions; under some training setups, they may also encourage base-rate policies. Market prices provide more frequent proxy targets but may contain liquidity, herding, and participant errors~\citep{levy2026eventforecasting,lee2025massivetraining}. Resolved questions may also appear in pretraining data. Studies should therefore document the knowledge cutoff and data provenance, and evaluate prospectively on questions that were unresolved during training (Section~\ref{ssec:contamination}).

The two reward sources therefore trade variance for bias. A binary resolution is authoritative but reveals little about which search or reasoning action helped, while a market trajectory supplies many intermediate targets whose prices may reflect liquidity, herding, or participant error. Training reports should identify which actions receive credit and whether reward density changes calibration rather than only accuracy.

\vspace{-0.1in} \paragraph{Dense and Calibration Rewards}
Dense step-wise and variable rewards can improve performance in some numerical forecasting models~\citep{liu2025timer1,zhang2025timemaster,li2026timerft}. Agentic systems can additionally use workflow-level or turn-level rewards to assign credit across a multi-stage reasoner--forecaster process~\citep{pan2026castflow,feng2026kairosagent}.
Proper-score theory explains why correctness alone does not penalize
overconfidence~\citep{gneiting2007strictly}. Empirically, adding Brier score to
correctness rewards improves accuracy and calibration
~\citep{damani2025beyond}. However, applying a reward to a single realized outcome can induce outcome-rationalizing reasoning. 

A post-hoc Beta--Bernoulli calibrator instead
uses a text encoder and mixture-valued output, and outperforms
forecasting-specific fine-tuning without imposing a monotone two-parameter
correction~\citep{dai2026betabernoulli}
(Section~\ref{ssec:uq}).
A possible leakage pathway is that gradients from a realized outcome encourage rationales that explain the observed label rather than beliefs supported by decision-time evidence. State-conditioned rewards and gradient masking may reduce this pathway, but finite-sample estimation still introduces noise when states are grouped by similarity. A compute-controlled comparison should therefore evaluate proper-score training and an external calibrator using matched data, model, and compute budgets. Otherwise, policy updates may receive credit for correcting an output distortion that a simpler calibrator could fix.

\subsection{Ablation Evidence}
\label{ssec:ablation}

Evidence on the incremental value of language-model components in time-series forecasting is mixed. One ablation study reports that replacing or removing the LLM component leaves accuracy unchanged or improves it, and that smaller models trained from scratch can outperform pretrained LLMs at lower compute~\citep{tan2024actually}. By contrast, a larger-scale re-evaluation reports positive language-model contributions and attributes earlier null results to limited evaluation scale~\citep{qiu2026rethinking}. Because these studies use different protocols, a decisive comparison should hold datasets, horizons, splits, preprocessing, tuning budgets, and random seeds fixed, and should report paired uncertainty intervals for the incremental language-model effect.

Such a comparison must also keep the surrounding scaffold fixed. Removing the LLM while changing encoders, normalization, preprocessing, or tuning budget evaluates a different system rather than the marginal contribution of language pretraining.

Recent critiques argue that aggregate rankings can obscure practical capability and decisive design choices such as global versus local modeling~\citep{phungtuaeng2026wronggame,moretti2025benchmarking}, reinforcing the need for leak-resistant benchmarks and contamination audits (Section~\ref{ssec:contamination})~\citep{li2026tsfmaudit}. To establish the incremental value of language components, systems should report ablations against purpose-built numerical forecasters, especially on univariate tasks~\citep{ansari2025chronos2,khwaja2026toto2}. LLM-specific training is most strongly motivated when the task requires unstructured context, genuinely novel events, or evaluated explanations for decision makers; these criteria should be defined and measured (Section~\ref{sec:benchmarks}).

%% file: sections/09_benchmarks.tex
Reliable evaluation requires leak-resistant, point-in-time benchmarks and reporting of appropriate forecast scores, uncertainty quality for probabilistic outputs, and inference costs. Numerical and event targets use different baselines, but both require predictions to depend only on information available before the decision time; benchmark construction must also audit pre-training contamination. Table~\ref{tab:benchmarks} compares temporal and contamination controls together with the primary metrics.

\begin{table}[t]
\centering
\scriptsize
\renewcommand{\arraystretch}{0.94}
\setlength{\tabcolsep}{2.2pt}
\caption{Forecasting benchmarks grouped into numerical, event, and market-settled tasks. Within each block, entries are ordered by the strength of their temporal and contamination controls. Returns and payoffs measure downstream decisions and are not substitutes for proper scores of predictive distributions.}
\label{tab:benchmarks}
\begin{tabularx}{\textwidth}{@{}>{\hsize=0.95\hsize\raggedright\arraybackslash}XW{0.80}W{1.05}W{1.20}W{1.00}@{}}
\toprule
Benchmark & Forecast task & Data / evidence source & Leakage / temporal control & Primary metric \\
\midrule
Monash archive~\citep{godahewa2021monash} & Numerical & Curated public series & Static archive; none & MASE, sMAPE \\
ProbTS~\citep{zhang2023probts} & Numerical; probabilistic & Standard benchmark suites & Static archive; none & CRPS, point error \\
Context is Key~\citep{williams2024context} & Numerical + text & Series + written context & Static archive; none & Region-weighted CRPS \\
GIFT-Eval~\citep{aksu2024gifteval} & Numerical; zero-shot & Static multi-domain archive & No universal training-overlap guarantee & MASE, CRPS \\
TSRBench~\citep{yu2026tsrbench} & Reasoning; multimodal & Multi-task series corpus & Fixed capability / task categories & Per-capability results \\
fev-bench~\citep{shchur2025fevbench} & Numerical; covariates & 100 real tasks; 46 with covariates & Rolling origins; held-out windows & Skill score; bootstrap CI \\
Dr-CiK~\citep{tang2026drcik} & Numerical + retrieval & Search-recovered context & Timestamped evidence pool & Region-weighted CRPS \\
TIME~\citep{qiao2026time} & Numerical; zero-shot & 50 fresh datasets; 98 tasks & Non-overlapping rolling windows; fresh-data control & MASE, CRPS; pattern / task ranks \\
\midrule
Autocast~\citep{zou2022autocast} & Event & Forecasting tournaments & Retrospective cutoff & Brier, accuracy \\
MIRAI~\citep{ye2024mirai} & Event; agentic & Relations event database & Temporal masking & Relation / event F1; distributions \\
WorldReasoner~\citep{chi2026worldreasoner} & Event + reasoning & Retrospective resolved events & Simulated dates; evidence / graph audit & Probability, evidence, graph \\
SocietyBench~\citep{wang2026societybench} & Event; counterfactual & Social-world scenarios & Anonymized entities + dates & Calibration + temporal accuracy \\
Hindcast~\citep{ye2026hindcast} & Event; replay & Resolved markets & Frozen pre-cutoff corpus & Brier / accuracy; market distance \\
Agentic Time Machine~\citep{chai2026timemachine} & Event; replay & Cutoff-filtered web reconstruction & Approximate historical web state & Offline replay score \\
Generated questions~\citep{bosse2026autoqgen} & Event & Agent-written and resolved & Generated after cutoff & Validity rate, Brier \\
ForecastBench~\citep{karger2024forecastbench} & Event & Markets, platforms, series & Unresolved at submission & Brier vs.\ human baseline \\
FutureX~\citep{zeng2025futurex} & Event; agentic & Live daily pipeline & Live; unresolved & Accuracy at resolution \\
ForecastBench-Sim~\citep{lee2026forecastbench} & Event; counterfactual & Simulated world & Unresolvable in reality & Simulator proper score \\
\midrule
TimeSeek~\citep{mostafa2026timeseek} & Event; market-settled & Market lifecycle snapshots & Lifecycle-stratified sampling & Skill by lifecycle stage \\
PolyBench~\citep{cheng2026polybench} & Event + simulated trading & Historical Polymarket order books & Timestamp-locked snapshots & Brier + simulated return \\
Prophet Arena~\citep{yang2025prophetarena} & Event; market-settled & Live market questions & Live; unresolved & Calibration and trading analyses \\
Prediction Arena~\citep{zhang2026predictionarena} & Event; market-settled & Live markets; real capital & Live; capital at risk & Realized profit / loss \\
Foresight Arena~\citep{nechepurenko2026foresight} & Event; proposed on-chain & Commit--reveal market design & Prospective results pending & Brier score; Alpha score \\
LiveTradeBench~\citep{yu2025livetradebench} & Trading & Live market state & Live & Return (not a proper score) \\
FinDeepForecast~\citep{li2026findeepforecast} & Event; financial & Continuously generated tasks & Live & Task-level accuracy \\
\bottomrule
\end{tabularx}
\end{table}

\subsection{Numerical and Event Benchmarks}
\label{sec:benchmarks}
\label{ssec:bench-event}

Monash, GIFT-Eval, and ProbTS support broad numerical comparisons, but static archives cannot guarantee absence from pre-training data
~\citep{godahewa2021monash,aksu2024gifteval,zhang2023probts}. FEV-Bench uses
rolling evaluation and bootstrapped skill scores across 100 real tasks, 46 of which include covariates~\citep{shchur2025fevbench}; TIME reports 50 recently collected datasets, 98 operationally aligned tasks, and non-overlapping rolling windows~\citep{qiao2026time}.
Data collected after a model's training period can reduce the risk of parametric overlap, whereas rolling-origin evaluation preserves deployment order and can reveal performance changes across time.These controls address different threats.

\vspace{-0.1in} \paragraph{Capability and Context Evaluation}
Specialized benchmarks isolate intermediate capabilities. TSRBench reports that increasing model scale improves perception and reasoning but not necessarily final prediction accuracy, whereas TFRBench evaluates reasoning traces separately from final forecasts; multi-turn protocols test consistency across repeated interactions~\citep{yu2026tsrbench,ahamed2026tfrbench,kong2026timesagemt}. Other benchmarks separate temporal order, spatial structure, magnitude, asset-pricing scenarios, or irregular-event handling~\citep{minnick2026spikeprophecy,wang2025fintsbridge,gupta2025last}. These decompositions matter because better perception or a more plausible trace does not imply a better forecast. Reporting only an aggregate score can hide a system that captures temporal order but fails on magnitude, horizon, or downstream action selection.
Multimodal benchmarks must verify that text adds predictive signal. LLM
forecasters often underuse context and that protocol choices can overstate the apparent gain
~\citep{williams2024context,liu2026timesx}. Under text collapse, arbitrary
filler produces the same boost because the context branch acts as a
content-independent transformation~\citep{nguyen2026textcollapse}. Scenario and
agentic benchmarks test context application and retrieval
~\citep{jang2026whatif,tang2026drcik}, but none isolate the incremental contribution of written context (Section~\ref{ssec:ablation}).

\vspace{-0.1in} \paragraph{Live, Market, and Replay Evaluation}
Event benchmarks compare systems with crowds and markets, where ensembling
drives substantial gains
~\citep{zou2022autocast,halawi2024approaching,schoenegger2024wisdom}. MIRAI
masks post-cutoff evidence to evaluate retrieval jointly with forecasting, ForecastBench reports that expert forecasters outperform the evaluated models, and FutureX documents the vulnerability of live retrieval to fabricated web pages
~\citep{ye2024mirai,karger2024forecastbench,zeng2025futurex}.

Market benchmarks differ in the realism of their decision setting. Prediction Arena exposes live capital to risk, whereas PolyBench simulates execution on historical, timestamp-locked order books and reports losses for many evaluated systems
~\citep{zhang2026predictionarena,cheng2026polybench}. Foresight Arena specifies an
on-chain commit--reveal design, but prospective results are not yet available~\citep{nechepurenko2026foresight}. Pooling early- and late-stage market questions can also confound the sampling strategy with forecasting ability
~\citep{mostafa2026timeseek}.
Replay benchmarks restore reproducibility through frozen corpora or reconstructed historical web states, but their agreement with live performance remains unknown
~\citep{ye2026hindcast,chai2026timemachine}. Automated pipelines can generate
questions and resolve them later with high reported validity, which can scale
live and replay evaluation~\citep{bosse2026autoqgen}.

\vspace{-0.1in} \paragraph{Rationale Auditing}
WorldReasoner and SocietyBench audit forecasts using evidence, causal graphs, or entity/date shifts~\citep{chi2026worldreasoner,wang2026societybench}. Simulated worlds create counterfactuals that are harder to memorize, but their transfer to real-world forecasting remains uncertain~\citep{lee2026forecastbench}. Accuracy often declines as evaluation data move further beyond the model's training period, making the documented cutoff an essential reporting item~\citep{dai2024prescient}. A system may outperform a general crowd while trailing expert forecasters, and different question distributions prevent direct cross-benchmark rankings~\citep{karger2024forecastbench}. Rationale audits test whether a forecast is supported by admissible evidence; they do not establish that the rationale caused the reported probability. Simulated and live evaluations therefore answer complementary questions: the former strengthens contamination control, while the latter preserves institutional cues and real base rates.

\subsection{Contamination and Leakage}
\label{ssec:contamination}

Having defined admissible forecasts under $\mathcal{I}_t$ and distinguished feature from parametric leakage (Section~\ref{ssec:formulation}), evaluation now relies on data audits, temporal partitioning, and outcome-independent consistency checks. This subsection compares the guarantees and limitations of leakage controls.

\vspace{-0.1in} \paragraph{Leakage Sources}
Feature leakage, where models use covariates unavailable at decision time, confounds model capability with information availability, necessitating strict point-in-time reconstruction of historical data including revisions~\citep{guan2026leakage}. Dataset reuse can also create train--test sample overlap and temporal overlap among correlated series, producing optimistic, non-generalizable evaluations~\citep{meyer2025leakage}.
Parametric leakage embeds benchmark data directly into model weights, rendering dataset cleaning insufficient. TSFMAudit reveals that text-adapted loss detectors are unreliable for time-series models, as low forecast loss often reflects series smoothness or strong generalization rather than memorization~\citep{li2026tsfmaudit}. Consequently, large-scale audits verifying zero-shot claims on standard archives remain unavailable.
This is important negative evidence: a detector failure does not prove contamination, but it also prevents a clean zero-shot guarantee. Fresh post-cutoff data and prospective tasks remain stronger controls than inferring memorization from unusually low loss.

\vspace{-0.1in} \paragraph{Leakage-Resistant Evaluation}
Date-masked retrospective evaluation fixes the information cutoff to approximate decision-time context, although residual parametric-leakage risks remain~\citep{ma2026oracleproto}. Live benchmarks reduce outcome memorization by testing unresolved questions, but each item expires at resolution (Section~\ref{ssec:bench-event}). Replay against frozen or cutoff-filtered corpora can block post-decision retrieval while preserving offline reproducibility; its agreement with live performance still requires direct validation~\citep{ye2026hindcast,chai2026timemachine}. Studies should specify the evaluation design and enumerate the leakage vectors that remain. Consistency checking evaluates logical and arithmetic constraints across a set of forecasts before resolution, providing an outcome-independent coherence diagnostic~\citep{paleka2024consistency}. The evidence links incoherence with higher resolution-based error, but coherence should be treated as a necessary diagnostic at most, not a sufficient ranking criterion. Benchmark shortcuts can also confound reasoning claims with retrieval or memorized domain knowledge~\citep{yang2024critical}. Cutoff contamination and diversity-based aggregation gains do not by themselves prove overlap in pre-training corpora~\citep{douven2026wisdomllmcrowds}. A 12-paper pilot audit reports scaffold and cost-disclosure gaps, motivating shared live infrastructure~\citep{moghadasi2026what}.

Figure~\ref{fig:benchmarks} maps representative benchmarks by contamination
control and decision relevance. Its decision-critical numerical region remains
empty: FEV-Bench improves ranking precision and TIME uses fresh data, but both
remain offline~\citep{shchur2025fevbench,qiao2026time}. Numerical evaluation
could adopt frozen-corpus replay, post-cutoff generation, or market settlement.
\input{figures/fig_benchmarks}
The map separates freshness from decision relevance: strong temporal controls
do not guarantee deployment realism, while market settlement does not supply a
numerical baseline. Evaluations should report both dimensions.
The empty region is therefore a research gap rather than a claim that live markets are universally superior. A benchmark enters that region only when it combines point-in-time numerical inputs, realistic downstream decisions, and a specialist numerical comparator under the same protocol.

\subsection{Scoring and Uncertainty}
\label{ssec:uq}

MAE and RMSE can rank models differently, while composite criteria add tail
sensitivity~\citep{gonzalez2025hierarchical}. Predictive distributions require
proper scores for calibration and sharpness~\citep{gneiting2007strictly};
PolyBench's overconfident losses illustrate the need
~\citep{cheng2026polybench}. Evaluations must also report inference costs,
especially for deep ensembles~\citep{kamarthi2021when}.

Proper scores penalize misplaced confidence and diffuse predictions together; calibration alone is insufficient because an uninformative base-rate forecast can be calibrated but not sharp.

\vspace{-0.1in} \paragraph{Confidence Elicitation}
Closed APIs typically expose only verbalized confidence, which correlates weakly with accuracy and underperforms output sampling~\citep{xiong2023can}. Alternative uncertainty measures encode different assumptions: sample entropy neglects unobserved sequence probability~\citep{kunitomojacquin2025role}, while semantic-clustering methods such as CSS depend on the chosen embedding representation~\citep{ao2024css}. Multi-forecast aggregation (Section~\ref{ssec:aggregation}) addresses errors that no single-model uncertainty estimate can reliably quantify.

On prediction markets, frontier models report probabilities exceeding observed frequencies, with reasoning-enhanced variants exhibiting higher expected calibration error due to extended deliberation inflating confidence without new evidence~\citep{nel2025kalshibench}. This mirrors the high-confidence financial losses seen in live market benchmarks (Section~\ref{ssec:bench-event})~\citep{cheng2026polybench}; Section~\ref{ssec:overconfidence} details this broader overconfidence issue.

Linear probing of open-weight model activations yields better-calibrated probabilities than verbalized forecasts, and many forecasts are largely determined before reasoning tokens are generated~\citep{sarfati2026probing}. This temporal separation indicates a faithfulness gap without establishing that every trace merely rationalizes a fixed answer. Models also update conservatively and inconsistently when exposed to post-cutoff evidence~\citep{yuan2025evolvecast}, revealing failures that static calibration tests miss.

\vspace{-0.1in} \paragraph{Correction and Task-Specific Uncertainty}
The learned Beta--Bernoulli calibrator (Section~\ref{ssec:train-calibration}) uses a text encoder and mixture-valued output and can outperform forecasting-specific fine-tuning~\citep{dai2026betabernoulli}. Proper-score training is another option~\citep{damani2025beyond}; \citet{singh2026verifiable} show that state-conditioned empirical-rate rewards and gradient masking are safer than naive single-outcome calibration rewards. Conformal prediction offers finite-sample marginal coverage under exchangeability~\citep{angelopoulos2021conformal}, with reported coverage gains for extreme AI weather forecasts and annual totals~\citep{asch2026rigorous}.
Marginal guarantees still do not ensure conditional coverage during rare tail events.

A review of 40 uncertainty quantification methods highlights limited ecological validity, an overemphasis on epistemic uncertainty, and the need for human-collaborative representations alongside calibration metrics~\citep{devic2025from}. Climate evidence instead distinguishes epistemic ensemble disagreement from aleatoric uncertainty: the former tracks shift and error, whereas the latter becomes unreliable under shift~\citep{mcafee2025confused}. Scalar confidence also discards spatial dependencies critical to physical forecasting (Section~\ref{ssec:app-physical}).

\subsection{Aggregation and Human--Agent Hybrids}
\label{ssec:aggregation}

Aggregation is central to human forecasting~\citep{tetlock2015superforecasting}.
On event targets, LLM ensembles can approach human crowd performance even when
individual models lag~\citep{schoenegger2024wisdom}. These gains depend on error
diversity alongside individual skill, because a weaker forecaster with
uncorrelated errors can add more value than a stronger but redundant model
~\citep{aitchison2026diversity}. Learned aggregators can therefore outperform
all pool members by exploiting disagreement signals
~\citep{douven2026wisdomllmcrowds}.

\vspace{-0.1in} \paragraph{Diversity and Deliberation}
Cross-model deliberation improves accuracy in heterogeneous groups but not homogeneous same-model groups, although the study does not directly measure error-pattern distinctness as the mechanism~\citep{schneider2025deliberating}. Assigning unique evidence subsets likewise avoids herding and encourages substantive revisions compared with sharing identical evidence~\citep{li2026infodelphi}. Multi-agent evaluations should report both model diversity and evidence overlap (Section~\ref{ssec:multi-agent}).

\vspace{-0.1in} \paragraph{Human and Architectural Aggregation}
A preregistered study shows that AI assistance improves human forecast accuracy even when the assistant is prompted to be overconfident, but it does not isolate the mechanism behind the gain~\citep{schoenegger2024aiaugmented}. A preliminary human--AI pilot associates collaborative traits with joint performance, so interaction design should be treated as an evaluation variable rather than a universal determinant~\citep{ming2026humancapital}.  Valid baselines must include both general crowds and expert superforecasters (Section~\ref{ssec:bench-event})~\citep{karger2024forecastbench}.
\citet{alur2025aiaforecaster} match superforecaster benchmarks by combining subquestion search, supervisor reconciliation, and calibration. Other studies evaluate argumentative coherence within one prediction and probabilistic consistency across related questions; these are distinct properties and should not be conflated with numerical aggregation~\citep{gorur2025argcoherent,paleka2024consistency}.

\subsection{Cost}
\label{sec:metrics}

Evaluations should separate development cost from per-query cost and report accuracy together with model calls, tokens, and latency. FLAIRR-TS improves accuracy through repeated refinement~\citep{jalori2025flairr}; however, multi-agent topology changes latency, call consolidation can discard tool information, and compute budgets are not directly comparable across different models~\citep{orogat2026understanding,ray2026agent,cheng2026model}. At minimum, reports should include calls per prediction, input and output tokens, wall-clock latency, and the strongest low-cost baseline at the same operating point. Some time-series foundation models support efficient zero-shot forecasting through small architectures or sparse mixture-of-experts routing~\citep{ekambaram2024ttm,shi2025timemoe}. Small models can also match or outperform LLM forecasters at lower compute on pure numerical tasks~\citep{tan2024actually}, while conformal prediction provides marginal post-hoc coverage guarantees only under its stated assumptions~\citep{angelopoulos2021conformal}. Evaluations should therefore report a cost--accuracy frontier and include explicit non-reasoning baselines whenever unstructured evidence is absent.

\vspace{-0.1in} \paragraph{Deployment and Distillation}
A renewable-energy review identifies latency, edge-hardware, and interoperability constraints but does not empirically establish tiered routing~\citep{manjunath2026llm}. Because tool value depends on the model and workload, these results motivate adaptive invocation rather than a fixed tool policy~\citep{cheng2026model}. Distillation can reduce inference cost, but the accounting must include the cost of generating teacher trajectories~\citep{li2026distilts,kang2025distilling}.

%% file: figures/fig_benchmarks.tex
\begin{figure}[t]
\centering
\ifdynamicfigures
  \begingroup
  \resizebox{\textwidth}{!}{%
    \input{figures/src/benchmarks.tex}%
  }
  \endgroup
\else
  \includegraphics[width=\textwidth]{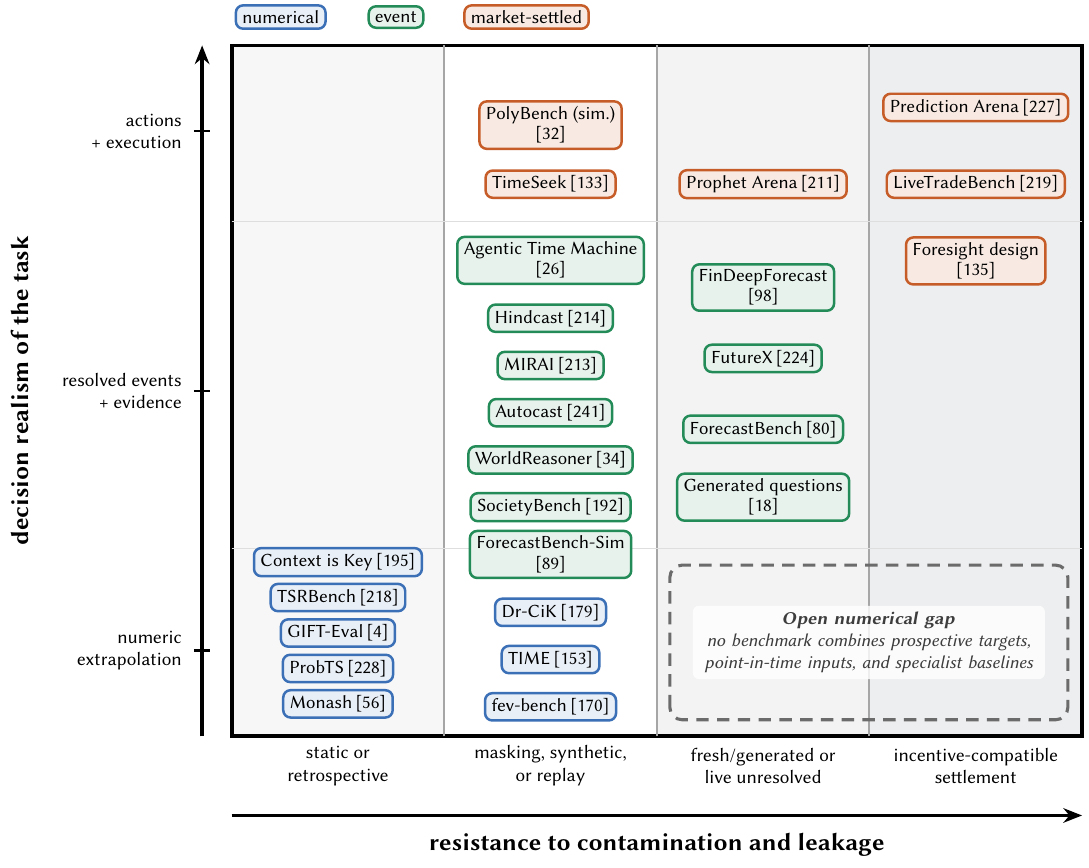}
\fi
\caption{All forecasting benchmarks in Table~\ref{tab:benchmarks}, positioned
by documented contamination and leakage control and decision relevance.
Coordinates summarize reported design properties, not measured rankings.
PolyBench uses simulated execution; Foresight Arena is a proposed design
without prospective results.}
\label{fig:benchmarks}
\end{figure}

%% file: figures/src/benchmarks.tex
\definecolor{cBenchNum}{HTML}{3B6FB6}%
\definecolor{cBenchEvent}{HTML}{238B57}%
\definecolor{cBenchMarket}{HTML}{C65D28}%
\newcommand{\labelsize}{\fontsize{9.2}{10.8}\selectfont}%
\newcommand{\benchsize}{\fontsize{9.0}{10.2}\selectfont}%
\begin{tikzpicture}[font=\figfont,
    bnum/.style={draw=cBenchNum, fill=cBenchNum!11, line width=1.2pt, rounded corners=3.5pt,
                 font=\figfont\benchsize, text=black, inner xsep=3.1pt,
                 inner ysep=2.5pt, align=center},
    bevt/.style={draw=cBenchEvent, fill=cBenchEvent!11, line width=1.2pt, rounded corners=3.5pt,
                 font=\figfont\benchsize, text=black, inner xsep=3.1pt,
                 inner ysep=2.5pt, align=center},
    bmkt/.style={draw=cBenchMarket, fill=cBenchMarket!13, line width=1.2pt, rounded corners=3.5pt,
                 font=\figfont\benchsize, text=black, inner xsep=3.1pt,
                 inner ysep=2.5pt, align=center},
    bandlab/.style={font=\figfont\labelsize, text=black, align=center}]

\def\W{14.4}\def\H{11.7}%
\def\B{3.6}%

\fill[cMute!6]  (0,0)       rectangle (3.6,\H);
\fill[white]    (3.6,0)     rectangle (7.2,\H);
\fill[cMute!9]  (7.2,0)     rectangle (10.8,\H);
\fill[cMute!13] (10.8,0)    rectangle (\W,\H);
\draw[draw=black, line width=1.8pt] (0,0) rectangle (\W,\H);
\foreach \x in {3.6,7.2,10.8}
  \draw[draw=black!35, line width=1.0pt] (\x,0) -- (\x,\H);
\foreach \y in {3.18,8.72}
  \draw[draw=black!13, line width=0.8pt] (0,\y) -- (\W,\y);

\node[bandlab] at (1.80,-0.50) {static or\\retrospective};
\node[bandlab] at (5.40,-0.50) {masking, synthetic,\\or replay};
\node[bandlab] at (9.00,-0.50) {fresh/generated or\\live unresolved};
\node[bandlab] at (12.60,-0.50) {incentive-compatible\\settlement};

\draw[axisline] (0,-1.34) -- (\W,-1.34);
\node[anchor=north, font=\figfont\bfseries\large, text=black] at (\W/2,-1.52)
     {resistance to contamination and leakage};

\draw[axisline] (-0.50,0) -- (-0.50,\H);
\foreach \y in {1.45,5.85,10.25}
  \draw[draw=black, line width=1.2pt] (-0.64,\y) -- (-0.36,\y);
\node[bandlab, anchor=east, align=right] (yl1) at (-0.72,1.45)
     {numeric\\extrapolation};
\node[bandlab, anchor=east, align=right] (yl2) at (-0.72,5.85)
     {resolved events\\+ evidence};
\node[bandlab, anchor=east, align=right] (yl3) at (-0.72,10.25)
     {actions\\+ execution};
\node[rotate=90, anchor=south, font=\figfont\bfseries\large, text=black]
     at ([xshift=-0.32cm]yl2.west) {decision realism of the task};

\node[bnum] at (1.80,0.55) {Monash\,\cite{godahewa2021monash}};
\node[bnum] at (1.80,1.15) {ProbTS\,\cite{zhang2023probts}};
\node[bnum] at (1.80,1.75) {GIFT-Eval\,\cite{aksu2024gifteval}};
\node[bnum] at (1.80,2.35) {TSRBench\,\cite{yu2026tsrbench}};
\node[bnum] at (1.80,2.95) {Context is Key\,\cite{williams2024context}};

\node[bnum] at (5.40,0.50) {fev-bench\,\cite{shchur2025fevbench}};
\node[bnum] at (5.40,1.30) {TIME\,\cite{qiao2026time}};
\node[bnum] at (5.40,2.10) {Dr-CiK\,\cite{tang2026drcik}};
\node[bevt] at (5.40,3.08) {ForecastBench-Sim\\[-0.5pt]\cite{lee2026forecastbench}};
\node[bevt] at (5.40,3.88) {SocietyBench\,\cite{wang2026societybench}};
\node[bevt] at (5.40,4.68) {WorldReasoner\,\cite{chi2026worldreasoner}};
\node[bevt] at (5.40,5.48) {Autocast\,\cite{zou2022autocast}};
\node[bevt] at (5.40,6.28) {MIRAI\,\cite{ye2024mirai}};
\node[bevt] at (5.40,7.08) {Hindcast\,\cite{ye2026hindcast}};
\node[bevt] at (5.40,8.06) {Agentic Time Machine\\[-0.5pt]\cite{chai2026timemachine}};
\node[bmkt] at (5.40,9.35) {TimeSeek\,\cite{mostafa2026timeseek}};
\node[bmkt] at (5.40,10.35) {PolyBench (sim.)\\[-0.5pt]\cite{cheng2026polybench}};

\node[bevt] at (9.00,4.05) {Generated questions\\[-0.5pt]\cite{bosse2026autoqgen}};
\node[bevt] at (9.00,5.20) {ForecastBench\,\cite{karger2024forecastbench}};
\node[bevt] at (9.00,6.40) {FutureX\,\cite{zeng2025futurex}};
\node[bevt] at (9.00,7.60) {FinDeepForecast\\[-0.5pt]\cite{li2026findeepforecast}};
\node[bmkt] at (9.00,9.35) {Prophet Arena\,\cite{yang2025prophetarena}};

\node[bmkt] at (12.60,8.05) {Foresight design\\[-0.5pt]\cite{nechepurenko2026foresight}};
\node[bmkt] at (12.60,9.35) {LiveTradeBench\,\cite{yu2025livetradebench}};
\node[bmkt] at (12.60,10.65) {Prediction Arena\,\cite{zhang2026predictionarena}};

\draw[draw=black!58, dash pattern=on 6pt off 4pt, line width=1.5pt,
      rounded corners=6pt] (7.42,0.28) rectangle (14.16,2.90);
\node[align=center, font=\figfont\fontsize{8.8}{10.6}\selectfont\itshape,
      text=black!76, text width=5.75cm, fill=white, fill opacity=0.82,
      text opacity=1, rounded corners=3pt, inner sep=3pt] at (10.79,1.59)
     {\textbf{Open numerical gap}\\
      no benchmark combines prospective targets,\\
      point-in-time inputs, and specialist baselines};

\node[bnum, anchor=west] at (0.05,\H+0.48) {numerical};
\node[bevt, anchor=west] at (2.30,\H+0.48) {event};
\node[bmkt, anchor=west] at (3.92,\H+0.48) {market-settled};

\end{tikzpicture}

%% file: sections/11_applications.tex
Application domains impose different decision objectives and evaluation constraints (Table~\ref{tab:applications}). In finance, forecast quality must be separated from portfolio returns; weather requires comparison with specialist numerical models; health forecasting requires prospective evaluation and careful handling of provisional data; and energy and operations add real-time integration constraints. Across these domains, LLMs are primarily used for unstructured context, workflow control, and explanation, while evidence that they improve numerical extrapolation remains limited.

\begin{table}[t]
\centering
\scriptsize
\setlength{\tabcolsep}{2.3pt}
\renewcommand{\arraystretch}{0.93}
\caption{Forecasting systems by application and evaluation.The bottom block contains specialized non-LLM baselines. In the cited studies, no LLM-based system has been shown to surpass these baselines on the primary weather or epidemiological forecast targets listed here (Section~\ref{ssec:app-physical}).}
\label{tab:applications}
\begin{tabularx}{\textwidth}{@{}>{\hsize=1.05\hsize\raggedright\arraybackslash}XW{0.85}W{1.20}W{0.90}W{1.00}@{}}
\toprule
System / study & Forecast domain & Core method & Inputs & Evaluation setting \\
\midrule
FinMem~\citep{yu2024finmem} & Equity trading & Memory hierarchy + risk persona & News + prices & Historical backtest \\
FinAgent~\citep{zhang2024finagent} & Multi-asset trading & Tool-augmented multimodal agent & Text + prices + charts & Historical backtest \\
FinCon~\citep{yu2024fincon} & Portfolio management & Conceptual-verbal reinforcement & Filings + news + prices & Historical backtest \\
FinPos~\citep{liu2025finpos} & Equity trading & Position-aware policy & Prices + text & Historical-market backtest \\
SIREN~\citep{ni2026siren} & Extreme-weather warning & Experience-grounded agents & Text + sensors & Warning-to-action chain \\
PandemicLLM~\citep{du2024pandemicllm} & Epidemic forecasting & LLM over multimodal signals & Policy + genomic + counts & Retrospective historical periods \\
AgentRx~\citep{jorf2026agentrx} & Clinical prediction & Multimodal agent benchmark & Records + images + notes & Held-out clinical tasks \\
NSW-EPNews~\citep{bi2025nsw} & Electricity price & News-augmented benchmark & Prices + news & Multi-step-ahead backtest \\
EventCast~\citep{hu2026eventcast} & E-commerce demand & Event knowledge + numerical model & Sales + event text & Retrospective backtest \\
\citet{liao2026bridging} & Air-ticket demand & Agentic context revision & Series + business context & Retrospective case studies \\
\midrule
GraphCast~\citep{lam2023graphcast} & Global weather & Graph network on reanalysis & Gridded fields & Deterministic verification \\
GenCast~\citep{price2024gencast} & Global weather & Diffusion ensemble & Gridded fields & Probabilistic verification \\
Aurora~\citep{bodnar2024aurora} & Earth system & Pre-trained; task fine-tuning & Multi-domain fields & Multi-task verification \\
COVID-19 Hub~\citep{cramer2022covidhub} & Epidemic forecasting & Multi-team quantile ensemble & Surveillance counts & Prospective weekly score \\
EPIFNP~\citep{kamarthi2021when} & Epidemic forecasting & Functional neural process & Surveillance counts & Real-time flu evaluation \\
\citet{mahmud2025hybrid} & Epidemic forecasting & Hybrid ARIMA--LSTM & Case counts & Retrospective split \\
\bottomrule
\end{tabularx}
\end{table}

\subsection{Finance and Prediction Markets}

Financial agents use LLMs for memory, multimodal perception, deliberation, and explanation. FinMem uses multi-timescale memory and explicit risk settings, whereas FinAgent processes charts, news, and prices~\citep{yu2024finmem,zhang2024finagent}. FinCon uses language-based feedback, TradingAgents simulates role-based agent interactions, and ElliottAgents adds technical-analysis constraints~\citep{yu2024fincon,xiao2024tradingagents,chudziak2025elliottagents}. Other systems evaluate explanations together with classification or portfolio outcomes, or modularize financial-agent capabilities~\citep{koa2024explainable,zhou2024finrobot}.

Directional price predictions do not directly determine trading positions. FinPos and FinRS evaluate policies that map forecasts to positions and model continuous position management and multi-timescale risk, rather than isolated trades~\citep{liu2025finpos,liu2025finrs}. Studies should therefore report forecast quality separately from position-dependent returns. This distinction is operational as well as metric-based: identical probabilities can produce different returns under different position sizing, turnover, and risk limits. 

Hybrid financial models assign narrower roles to LLMs, such as extracting directional sentiment, confidence, or asset-dependence features~\citep{hussain2026improving,yi2026paradigm}. In regime-conditioned volatility forecasting, a frozen LLM directly predicts realized variance and outperforms the reported classical baselines, but the study does not isolate which component produces the gain~\citep{asaad2026regime}. Existing evaluations therefore do not establish whether workflow agents or feature-based hybrids provide higher accuracy at matched cost.

Market-linked evidence is less favorable: PolyBench reports simulated losses on historical order-book snapshots~\citep{cheng2026polybench}, while LiveTradeBench evaluates live market states~\citep{yu2025livetradebench}. M6 likewise documents a gap between forecasting and investment performance~\citep{makridakis2024m6}. Because returns conflate predictive skill with execution and position sizing, evaluations should separately report proper probability scores, portfolio returns, passive baselines, and execution costs. InvestorBench emphasizes decision quality rather than a separate proper forecast score~\citep{li2024investorbench}, whereas FinDeepForecast supports prospective task evaluation~\citep{li2026findeepforecast}.

\subsection{Weather, Energy, Others}
\label{ssec:app-physical}

Specialized AI weather models provide strong numerical baselines. FourCastNet matches operational IFS at short leads for large-scale variables and exceeds it for selected fine-scale variables~\citep{pathak2022fourcastnet}; Pangu-Weather, GraphCast, GenCast, and Aurora report broader gains on their evaluated targets~\citep{bi2022pangu,lam2023graphcast,price2024gencast,bodnar2024aurora}. No LLM agent matches these systems in numerical prediction; weather agents such as SIREN and DORA instead support downstream response workflows~\citep{ni2026siren,wang2026can}.
For these agents, the relevant comparison is not an LLM-generated weather field against a specialist numerical model. It is whether language-based coordination improves warning interpretation, tool use, jurisdiction selection, or response timing while the numerical forecast is held fixed. Such workflow gains should not be reported as improved meteorological skill.

In energy forecasting, time-aligned news and quantile/conformal calibration complement numerical baselines~\citep{bi2025nsw,huang2025llm}; a renewable-energy review identifies latency and edge constraints that motivate, but do not empirically prove, selective LLM routing~\citep{manjunath2026llm}. EventCast incorporates promotions and holidays through an event database and LLM summaries for e-commerce demand~\citep{hu2026eventcast}, while \citet{liao2026bridging} revise air-ticket forecasts with business context in retrospective case studies. Evaluations should account for prediction-interval variance~\citep{lebedev2025analyzing} and measure incremental agent value over raw model outputs under domain-relevant costs.

%% file: sections/12_risks.tex
LLM forecasting introduces overconfidence, tool-access, and reflexivity risks.
Table~\ref{tab:failures} distinguishes failures that primarily degrade forecasting performance from evaluation pathologies that can inflate apparent performance or mask miscalibration. The following discussion uses this distinction to motivate priorities for measurement, training, and theory.

\begin{table}[t]
\centering
\scriptsize
\setlength{\tabcolsep}{2.4pt}
\renewcommand{\arraystretch}{0.96}
\caption{Documented failure modes by component. The first group lowers observed performance; the second can inflate scores or understate uncertainty and requires calibration, leakage, or reporting audits (Section~\ref{ssec:contamination}).}
\label{tab:failures}
\begin{tabularx}{\textwidth}{@{}>{\hsize=0.40\hsize\raggedright\arraybackslash}X>{\hsize=1.60\hsize\raggedright\arraybackslash}X@{}}
\toprule
Component & Documented failure modes \\
\midrule
\multicolumn{2}{@{}l}{\itshape Observed as lower forecast or task performance} \\
\addlinespace[1pt]
Representation & Number fragmentation~\citep{yang2025tokon}; ignored predictive context~\citep{williams2024context}; merged-call information loss~\citep{ray2026agent} \\
Temporal & Failed zero-shot reasoning~\citep{merrill2024struggle}; in-context limits~\citep{zhou2025why}; temporal-validity violations~\citep{zeng2025futurex}; retrieval under shift~\citep{zhou2026semantics} \\
Reasoning & Reasoning--answer mismatch~\citep{wang2026doing}; knowing--doing tool gap~\citep{cheng2026model}; perturbation brittleness~\citep{park2025revisiting}; over-privileged tool choice~\citep{yang2026when} \\
\addlinespace[2pt]
\multicolumn{2}{@{}l}{\itshape Detected by calibration, leakage, or reporting audits} \\
\addlinespace[1pt]
Calibration & Unreliable verbal confidence~\citep{xiong2023can}; entropy omits unseen mass~\citep{kunitomojacquin2025role}; failed tail coverage~\citep{asch2026rigorous}; reasoning-induced miscalibration~\citep{nel2025kalshibench} \\
Systemic & Decision-time feature leakage~\citep{guan2026leakage}; perishable benchmark scores~\citep{gilda2026position}; retrieval-confounded shortcuts~\citep{yang2024critical}; sampled reporting gaps~\citep{moghadasi2026what} \\
\bottomrule
\end{tabularx}
\end{table}

\vspace{-0.1in} \paragraph{Overconfidence}
\label{ssec:overconfidence}

Fluent explanations can mask miscalibration: verbal confidence is not reliably aligned with accuracy, and trading-oriented full system show that high confidence can persist even when decisions incur losses~\citep{xiong2023can,cheng2026polybench}. Proper scoring rules and coverage tests are therefore necessary because surrogate uncertainty metrics do not necessarily track realized forecast reliability~\citep{devic2025from,camporeale2025verification}. Rationale length should not be treated as evidence of forecast quality or calibration. Retrospective benchmark results can become stale as models and data change, while apparent performance can also be inflated by decision-time leakage~\citep{gilda2026position,guan2026leakage}. Reporting audits additionally identify gaps in scaffold and cost disclosure, but disclosure quality alone does not establish forecast validity~\citep{moghadasi2026what}. 

\vspace{-0.1in} \paragraph{Tool access and reflexivity}
\label{ssec:tool-risk}

Broad tool permissions can turn a forecasting error into an unnecessary order, alert, or other downstream action~\citep{yang2026when}. Open-web retrieval can expose live forecasters to fabricated or low-quality evidence, whereas restricting retrieval to vetted corpora reduces this risk at the potential cost of timeliness~\citep{zeng2025futurex}. As agents gain access to more tools, the set of possible downstream actions grows, increasing the potential consequence of a forecasting error. In reflexive domains such as markets, published forecasts can influence the outcomes they are intended to predict. Because ensemble gains depend on both component skill and error diversity, deployment should monitor correlated failure modes rather than model accuracy alone~\citep{aitchison2026diversity}. Prospective market-linked benchmarks could test such feedback, but existing evaluations do not yet establish reflexive effects: some rely on historical simulation, while prospective evidence remains pending~\citep{cheng2026polybench,nechepurenko2026foresight}.

\vspace{-0.1in} \paragraph{Measurement priorities}

Unresolved or continuously refreshed questions, simulated environments, replay-based evaluation, and consistency checks address different forms of leakage, benchmark staleness, and evaluation instability~\citep{karger2024forecastbench,zeng2025futurex,lee2026forecastbench,ye2026hindcast,paleka2024consistency}. Point-in-time covariates strengthen temporal validity but do not by themselves prevent leakage through retrieval, benchmark construction, or subsequently revised labels. Because these mechanisms address different failure pathways, no single evaluation protocol is sufficient. Ablations should hold the surrounding scaffold fixed while removing or replacing the LLM component, and the full system should also be compared with zero-shot LLM, classical/statistical, and time-series foundation-model baselines~\citep{tan2024actually,qiu2026rethinking,ansari2025chronos2,hyndman2021forecasting,ansari2024chronos,das2024timesfm}. To test whether language inputs add value, evaluations should include settings in which predictive text contains information absent from the numerical inputs, which LLM systems can otherwise underuse~\citep{williams2024context}. Forecast scores should be reported alongside tool calls, token use, latency, and monetary cost, which remain inconsistently documented in reporting audits of agent-based forecasting~\citep{moghadasi2026what}. Because the value of tool use and the effects of excessive reasoning vary across models and tasks, routing policies require backbone- and workload-specific evaluation~\citep{cheng2026model,zhou2026overthinking}.

\vspace{-0.1in} \paragraph{Training and adaptation}

Proper-score training should be compared with learned post-hoc calibration, which can outperform forecasting-specific fine-tuning~\citep{gneiting2007strictly,dai2026betabernoulli}. Because naive single-outcome rewards can corrupt reasoning, comparisons should include state-conditioned, gradient-masked rewards on unresolved questions~\citep{singh2026verifiable}. One realized outcome can reward luck rather than a calibrated predictive belief. Under regime shifts, similarity retrieval degrades and regime-conditioned methods add state-estimation error~\citep{zhou2026semantics,asaad2026regime}. External memory and online residual adaptation avoid weight retraining~\citep{zhou2026externalization,dai2026orca}, but benchmarks must still measure break-detection delay, post-break error, and recovery.

\vspace{-0.1in} \paragraph{Shared protocols and collaboration}

Although both target types estimate predictive distributions, event protocols emphasize live evaluation and aggregation, while numerical protocols emphasize horizons, distributional metrics, and foundation-model baselines. A shared protocol for readouts~\eqref{eq:numerical} and~\eqref{eq:event} should define decision-time information and require proper scores. Common admissibility rules would permit comparison without erasing target-specific horizons or actions. Human--agent collaboration also needs direct tests. Assistance can improve human accuracy, AIA Forecaster matches one superforecaster benchmark, and a separate evaluation still places frontier LLMs below experts~\citep{schoenegger2024aiaugmented,alur2025aiaforecaster,lu2025expertforecasters}. A pilot links collaborative traits to joint performance~\citep{ming2026humancapital}; diverse-model deliberation helps where homogeneous groups do not, so evaluations should track diversity and ensemble performance~\citep{schneider2025deliberating}.

%% file: sections/14_conclusion.tex
Evidence for LLM-based forecasting agents remains mixed. Their most plausible role is where forecasting requires language evidence, event semantics, tool interaction, or auditable explanations. These capabilities can complement structured models, but do not establish LLMs as general replacements for statistical, numerical, or time-series foundation models. Text can provide useful predictive information, yet systems often underuse it~\citep{williams2024context}. Controlled ablations are therefore important because gains may arise from retrieval, prompts, interfaces, aggregation, or numerical backbones rather than the LLM itself; removing the language component can leave accuracy unchanged or improved~\citep{tan2024actually}.

Current conclusions are limited by decision-time leakage, pre-training contamination, benchmark staleness, incomplete calibration and cost reporting, and reliance on retrospective evaluation~\citep{guan2026leakage,gilda2026position,moghadasi2026what}. ForecastBench uses unresolved questions and finds experts ahead of evaluated models, while PolyBench reports simulated execution losses rather than live-capital evidence~\citep{karger2024forecastbench,cheng2026polybench}. Evaluation should therefore consider proper scores, uncertainty quality, inference cost, and decision relevance alongside predictive accuracy.

Studies should use live or time-stamped evidence, train and evaluate probabilities with proper scores and calibration-aware objectives~\citep{gneiting2007strictly,singh2026verifiable}, and report component-level ablations against strong non-LLM baselines. Model calls, tokens, latency, and development costs should accompany predictive scores. Retrieved evidence and tool actions should have auditable provenance, while deployment should monitor distribution shift, correlated errors, and forecast-induced feedback. Hybrid systems should be used only when their semantic or interactive capabilities provide measurable decision value at acceptable cost. The central question is when LLM participation yields a reproducible improvement over simpler alternatives.